\documentclass{article} 
\usepackage{iclr2027_conference,times}

\usepackage[utf8]{inputenc}
\usepackage[T1]{fontenc}

\usepackage{subcaption}

\usepackage{graphicx}
\usepackage{url}
\usepackage{booktabs}
\usepackage{amsmath,amssymb,amsfonts,mathtools,amsthm}
\usepackage{nicefrac}
\usepackage{microtype}
\usepackage{xcolor}
\usepackage{algorithm}
\usepackage{algorithmic}
\usepackage{newfloat}
\usepackage{listings}
\usepackage{placeins}
\usepackage{multirow}
\usepackage{hyperref,longtable}
\usepackage[capitalize,noabbrev]{cleveref}
\newtheorem{theorem}{Theorem}
\newtheorem{lemma}{Lemma}
\newtheorem{proposition}{Proposition}
\newtheorem{corollary}{Corollary}
\newtheorem{remark}{Remark}

\newcommand{\inlineeq}[2]{%
  \refstepcounter{equation}%
  \label{#1}%
  {\small\ensuremath{\textstyle #2}}%
  \nobreak\hspace{0.2em}\mbox{(\theequation)}%
}
\let\OriginalIncludeGraphics\includegraphics
\renewcommand{\includegraphics}[2][]{%
  \IfFileExists{#2}{\OriginalIncludeGraphics[#1]{#2}}{%
    \fbox{\begin{minipage}[c][0.18\linewidth][c]{0.88\linewidth}\centering\footnotesize Missing figure file: {\ttfamily\detokenize{#2}}\end{minipage}}%
  }%
}

\iclrfinalcopy 

\title{Focus on Likely Classes for Test-Time Prediction}

\author{
  Johannes Schneider \\
  Department of Computer Science and Information Systems \\
  University of Liechtenstein \\
  Vaduz, Liechtenstein \\
  \texttt{johannes.schneider@uni.li} \\
}

\begin{document}

\maketitle
\lhead{Preprint.} 

\begin{abstract}
We ask: \textbf{\emph{Can focusing on likely classes of a single, in-domain sample improve accuracy? Prior work argued ``no'', we answer ``yes'' on average}}. Standard entropy minimization yields largest gains, and we further dissect why by looking at its two core mechanisms: increasing likely and decreasing unlikely class predictions. Simply maximizing logits of the two most likely classes explains most of its benefits and, in some cases, even outperforms entropy minimization. Thus, decreasing classes is a secondary concern. 

Our controlled experiment and theory investigate the most puzzling case, why simply optimizing the logits of the two most likely classes  leads to accuracy gains. Our small scale networks suggest that the degree of uncertainty of a prediction, model choice, and to a lesser extent dataset choice moderate the success chances of the optimization. We show that the true class tends to have larger gradients independent of whether the prediction is correct or not. Our theory thoroughly investigates one of multiple possible explanations from the angle of shared features. 

Our evaluation focuses on 26 large language models and 12 text datasets, and 18 image recognition models trained on ImageNet. It demonstrates gains of 0.6\% on average for entropy minimization on uncertain samples using no extra data aside from the given test sample using a fixed learning rate. We also suggest \emph{gradient-ray ensembling}. When traversing input/embedding space by applying the same gradient with differing learning rates and \textbf{\emph{aggregating probabilities for the same sample further improves accuracy}} significantly to 0.75\% relying on a coarse range for learning rates, thus improving accuracy and eliminating the need to find a precise learning rate at the expense of computation.
\end{abstract}

\section{Introduction} \label{sec:intro}
State-of-the-art optimization methods in supervised and self-supervised learning, prevalent across subdisciplines ranging from image recognition in computer vision to text generation with large language models (LLMs), often minimize cross-entropy loss. The optimization objective aligns with the ideal scenario where a model assigns probability one to the correct class and zero to all others, achieving perfect class discrimination -- even on random data \citep{zha21un}. However, during test-time, a model typically exhibits uncertainty, particularly in cases where it errs. Prior work showed that a prediction of a single sample can be improved using augmentations of that sample (e.g., MEMO \cite{zha22memo}). However, finding the right augmentations can be non-trivial and evaluating many augmentations per single sample is computationally costly. Most works focus on domain adaption with the goal of continual learning based on many samples with a distribution shift. The shared dominant objective (e.g.,\cite{wan20tent,hu25pals,niu23sar,hu25reca}) \emph{\textbf{entropy minimization was claimed to lead to a trivial solution for a single sample confirming just the most likely class \citep{wan20tent}}}. The claim seems reasonable as entropy minimization by definition just seems to make the most likely class more likely and even if it did not, it seems questionable that reducing entropy of just a single sample should fix errors more often than it causes them. In this work, we question the existing ``wisdom'' that optimizing a single sample leads to no gains. We provide compelling evidence against it. \emph{\textbf{We also dissect entropy into two mechanisms: Increasing likely classes and decreasing unlikely classes}}. The benefit of each is questionable: Classes beyond the most likely are known to provide valuable signals \cite{hinton15distill,furlanello18born,zhao22dkd}. Thus, decrementing them might be harmful. Also, the benefits of incrementing likely classes are unclear. Assume one increments the logits of the two most likely classes with the same learning rate. In typical test data (and as we also show) the most likely class is more often correct than the second most likely. So naively switching them just increases the error, also incrementing both with the same learning rate should just leave their order in place.
Empirically, we show that each mechanism in isolation and their combination (entropy minimization) lead to gains on over 300 model-dataset pairs on average. Incrementing likely classes is clearly more beneficial than decrementing unlikely.  To better understand why incrementing classes leads to gains (and why not) we perform a controlled experiment and conduct a theoretical analysis. 

As the naive optimization method is computationally very expensive due to the need to conduct potentially tens of forward and backward passes per sample, we employ a number of computational optimizations, e.g., we do an uncertainty assessment step to limit our method to cases where it likely helps, we optimize only a small subset of parameters, i.e., token embeddings for text and we treat images as input parameters and optimize them.
Still, our most computationally efficient method still requires one additional forward and backward pass to deliver accuracy gains of 0.6\% and the identification of a model (but not dataset) dependent learning rate. However, we also identify a novel mechanism for self-ensembling, i.e. \emph{gradient-ray ensembling}. It uses input/embedding space traversal that further improves accuracy by using more compute and also makes the identification of a learning rate mostly obsolete by allowing a coarse grained range. It aggregates probability estimates of classes obtained by applying the same gradient with different learning rates and picks the most likely class of the aggregate.

Our contribution is to provide fundamental insights into the workings of entropy optimization for test-time adaptation that are also of immediate practical value and a mechanism to trade computation for accuracy gains and engineering effort based on gradient-ray ensembling. 


\begin{figure*}
\centering
\includegraphics[width=0.9\linewidth]{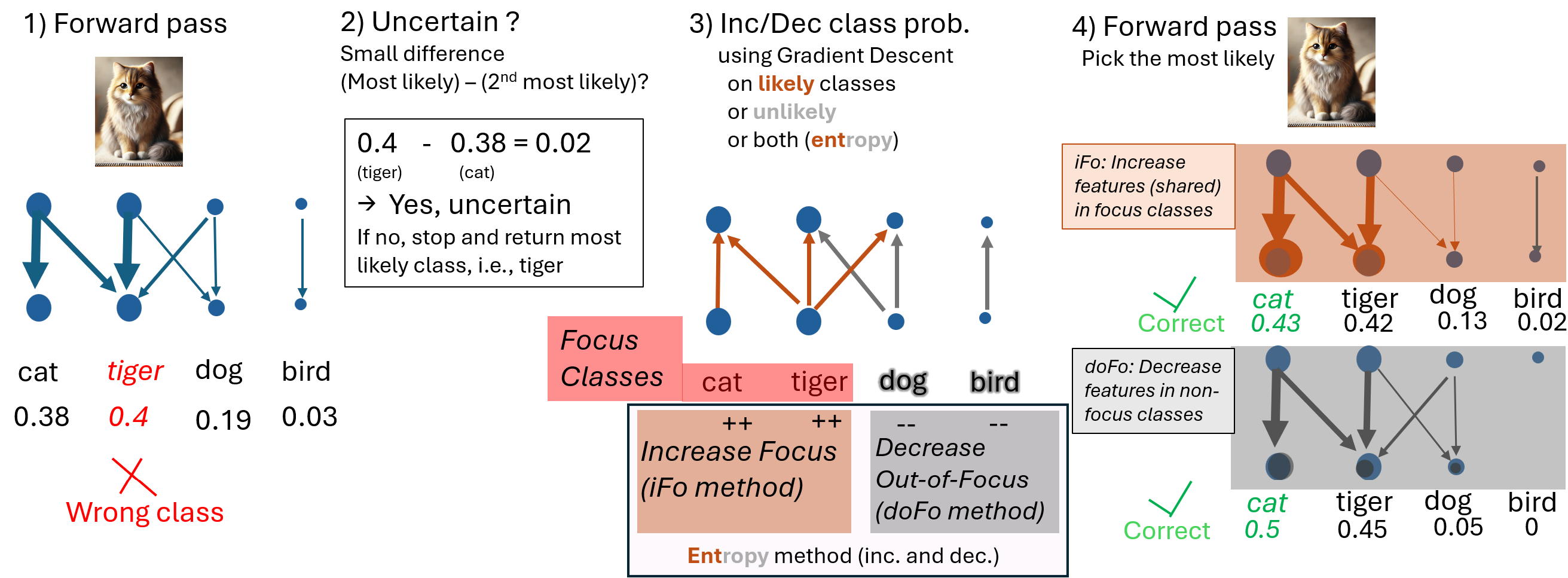}
\caption{Test-time approaches (iFo/doFo/entropy): If a prediction exhibits high uncertainty, changes are done to focus on likely classes either by increasing focus classes or decreasing all others or both.}
\label{fig:over}
\end{figure*}

\section{Methodology} \label{sec:metho}
Next, we describe how we aim to improve a prediction at test-time, as outlined in Figure \ref{fig:over}. We add two components to the classical (single-forward) prediction process: (i) An \emph{uncertainty assessment} procedure which decides whether to apply the focus optimization or not as gains are only probable on uncertain samples; (ii) A \emph{focus optimization} module, which alters predictions.

\textbf{Uncertainty Assessment: } The prediction uncertainty assessment determines whether focus optimization should be applied to a specific sample or not. On certain samples our method has a low probability of improving predictions as we also discuss later. We quantify uncertainty based on how strongly a classifier favors the predicted class compared to other classes. We measure the uncertainty as the difference between the probabilities of the most likely and second-most likely classes. If this difference is large, the prediction is intuitively likely correct, as alternative classes have significantly lower probabilities. Concretely, we consider a model $f$, such as a neural network, that outputs logit values $f_c(X)$ for each class $c \in C$. The softmax layer converts the logit values $f_c(X)$ into probability estimates $p_c(X)$. Let $m(i)$ denote the class with the $i$-th highest probability for input $X$; thus, $p_{m(1)}(X)$ and $p_{m(2)}(X)$ represent the highest and second-highest probabilities, respectively. We define the difference as:
\vspace{-3pt}
\begin{equation}
\Delta_{1,2}=p_{m(1)}(X)-p_{m(2)}(X)\label{eq:d12}
\end{equation}
Similar separations have been made for domain adaptation but with a performance goal rather than to save on computation, e.g., \cite{hu25pals}. 

%

\textbf{Focus Optimization:}
We employ gradient descent for a fixed number of iterations (often just 1) for the methods illustrated in Figure \ref{fig:over}: (i) \emph{Increasing outputs of Focus classes $F$ (iFo)}, (ii) \emph{Decreasing outputs of Out-of-Focus classes $C\setminus F$ (doFo)} ($C$ is the set of all classes), and (iii) doing both jointly (entropy). We do this for each sample anew (on the original network) as we require high learning rates that might negatively affect the network beyond the given sample.  Visually, in the center panel of Figure \ref{fig:over}, iFo increases activations of the subnetworks indicated by beige arrows, whereas doFo decreases those indicated by grey arrows. Entropy combines both ideas. 

Entropy minimization implicitly defines focus classes $F\subset C$ to be increased and non-focus classes $C\setminus F$ to be decreased. Entropy is \inlineeq{eq:entropy}{L^{Ent}(X)=-\sum_{c\in C}p_c(X)\log p_c(X)}, with softmax probabilities $p_c(X)$ defined as \inlineeq{eq:softmax}{p_c(X)=\frac{\exp(f_c(X))}{\sum_{k\in C}\exp(f_k(X))}}.

The gradient with respect to logit $f_c(X)$ is \inlineeq{eq:entropygrad}{\frac{\partial L^{Ent}(X)}{\partial f_c(X)}=p_c(X)\left(\bar f(X)-f_c(X)\right)}, where $\bar f(X)$ is the probability-weighted mean logit defined as \inlineeq{eq:meanlogit}{\bar f(X)=\sum_{k\in C}p_k(X)f_k(X)}.

Using the chain rule, the entropy gradient with respect to the model parameters $\theta$ is therefore
\begin{equation}
\small
\frac{\partial L^{Ent}(X)}{\partial\theta}
=
\sum_{c\in C}
p_c(X)\left(\bar f(X)-f_c(X)\right)
\frac{\partial f_c(X)}{\partial\theta}.
\label{eq:entropyupdateEnt}
\end{equation}
The term $p_c(X)\left(\bar f(X)-f_c(X)\right)$ acts as a scalar weight on the class-specific gradient $\frac{\partial f_c(X)}{\partial\theta}$. The weight is composed of two parts: the probability $p_c(X)$ regulates the scale of updates. In particular, without it unlikely classes with close to zero probability might receive substantial updates, which might hamper features. The second term has two functions: (i) it regulates scale by increasing updates for a class $c$ the further its logit $f_c(X)$ is from the mean $\bar f(X)$ and (ii) its sign $\operatorname{sign}\left(\bar f(X)-f_c(X)\right)$ implicitly defines whether a class belongs to the focus classes $c\in F$ or the non-focus classes. That is, \inlineeq{eq:entropyfocus}{F^{Ent}=\left\{c\in C\mid f_c(X)>\bar f(X)\right\}}. Applying $\log$ in Eq.~\ref{eq:softmax} to obtain $\log p_c(X)=f_c(X)-\log\sum_k\exp(f_k(X))$, we can write \inlineeq{eq:entropylogit}{f_c(X)-\bar f(X)=\log p_c(X)+L^{Ent}(X)}, and therefore \inlineeq{eq:entropyprobthreshold}{c\in F^{Ent}\Longleftrightarrow p_c(X)>\exp\left(-L^{Ent}(X)\right)}.

We also consider an explicit definition of $F^{k}$ with $|F|=k=2$, where $F$ consists of the $k=2$ most likely classes for a sample $X$. In this version, we drop the term $\left(\bar f(X)-f_c(X)\right)$, which implicitly defines $F$ and, in our theoretical derivation, even $p_c(X)$. The motivation is two-fold: (i) entropy minimization gives zero gradient and thus has no impact in the most relevant obvious case of maximum uncertainty, e.g., if two classes are equally likely and all others have zero probability, i.e., $p_0=p_1=0.5$; and (ii) this provides a simple version for understanding underlying mechanisms using a controlled experiment and theory. 

Thus, we obtain the gradients for iFo and doFo with $F \in \{F^{Ent},F^{k}\}$ based on Eq.~\ref{eq:entropygrad}:
\begin{equation}
\small
\frac{\partial L^{iFo}(X)}{\partial\theta}
=
\sum_{c\in F}
p_c(X)\left(\bar f(X)-f_c(X)\right)
\frac{\partial f_c(X)}{\partial\theta}.
\label{eq:entropyupdateiFo}
\end{equation}
and
\begin{equation}
\small
\frac{\partial L^{doFo}(X)}{\partial\theta}
=
\sum_{c\in C\setminus F}
p_c(X)\left(\bar f(X)-f_c(X)\right)
\frac{\partial f_c(X)}{\partial\theta}.
\label{eq:entropyupdateDoFo}
\end{equation}

\begin{algorithm}[tb]
\caption{Focus on the Likely: iFo, doFo, and Ent(ropy)}
\label{alg:focusopt}
\begin{algorithmic}[1]
\REQUIRE $f(\cdot|\theta)$: Model with parameters $\theta$; $X$: Input;\\
$Z^{Opt}\in\{X,\theta_{\mathrm{emb}},\theta\}$: Parameters to optimize (default: images: inputs $X$ ; text: token embeddings $\theta_{\mathrm{emb}}$ )\\ 
$d_{1,2}$: Uncertainty threshold (default: 0.05, but mostly implicitly defined using $d_{1,2,\%}$;);\\
Opt $\in \{\text{Ent},\text{iFo},\text{iFo}^k,\text{doFo}\}$ (default: Ent);\\
($\eta$,$T$): Initial learning rate $\eta$ and Iterations $T$  (default: Aggregating probabilities: text (1e-6, 13), images (1e-6/3,20); fixed lr:  model-dependent (see Appendix);\\

\STATE Compute logits $f(X)=[f_c(X)]$ and probabilities
$p^0_c(X)=\frac{e^{f_c(X)}}{\sum_{j\in C}e^{f_j(X)}}$ for $c\in C$
\STATE Sort probabilities so that $p^0_{m(1)}(X)\ge p^0_{m(2)}(X)\ge\ldots$
\STATE Compute $\Delta_{1,2}=p^0_{m(1)}(X)-p^0_{m(2)}(X)$ \COMMENT{Eq. (\ref{eq:d12})}
\STATE \textbf{if} {$\Delta_{1,2}\ge d_{1,2}$} \textbf{then} return \textbf{end if} \COMMENT{Uncertain Sample?}
\STATE \textbf{if} {Opt $=\text{iFo}^k$} \textbf{then} $F\leftarrow\{m(i)\mid i\in[1,k]\}$ \COMMENT{Optimize only Top-k}
\STATE \textbf{else} \COMMENT{Optimize increment (dynamic) top-k or decrement non-top-k or both (=entropy)}
\STATE \qquad Compute $L^{Ent}(X)=-\sum_{c\in C}p_c(X)\log p_c(X)$
\STATE \qquad $F:=\{c\in C\mid p_c(X)>\exp(-L^{Ent}(X))\}$ \COMMENT{Eq. (\ref{eq:entropyprobthreshold})} 
\STATE \textbf{end if}
\STATE $\nabla Z$=$\nabla_{Z^{Opt}}L^{Opt}(X)$ \COMMENT{Compute gradients for $Z_{Opt}$ once}
\STATE \textbf{for} {$t=0$ to $T-1$} \textbf{do} 
\STATE \quad $Z^{Opt}\leftarrow Z^{Opt}-\eta e^t \nabla Z$ \COMMENT{Update $Z_{Opt}$}
\STATE \quad Compute $p^{t+1}_c$ using $f$ with $Z^{Opt}$ \COMMENT{Reevaluate $f$ with changed parameters or inputs $Z_{Opt}$}
\STATE \textbf{return} $\arg\max_{c\in C} p^{T}_{c}(X)$ and $\arg\max_{c\in C} (\sum_i p^i_c(X))$  \COMMENT{Return most likely class for last iteration and aggregate}
\end{algorithmic}
\end{algorithm}

\section{Experiments on real-world data} \label{sec:exper}

\begin{figure}[t]
\centering
\vspace{-6pt}
\includegraphics[width=\linewidth]{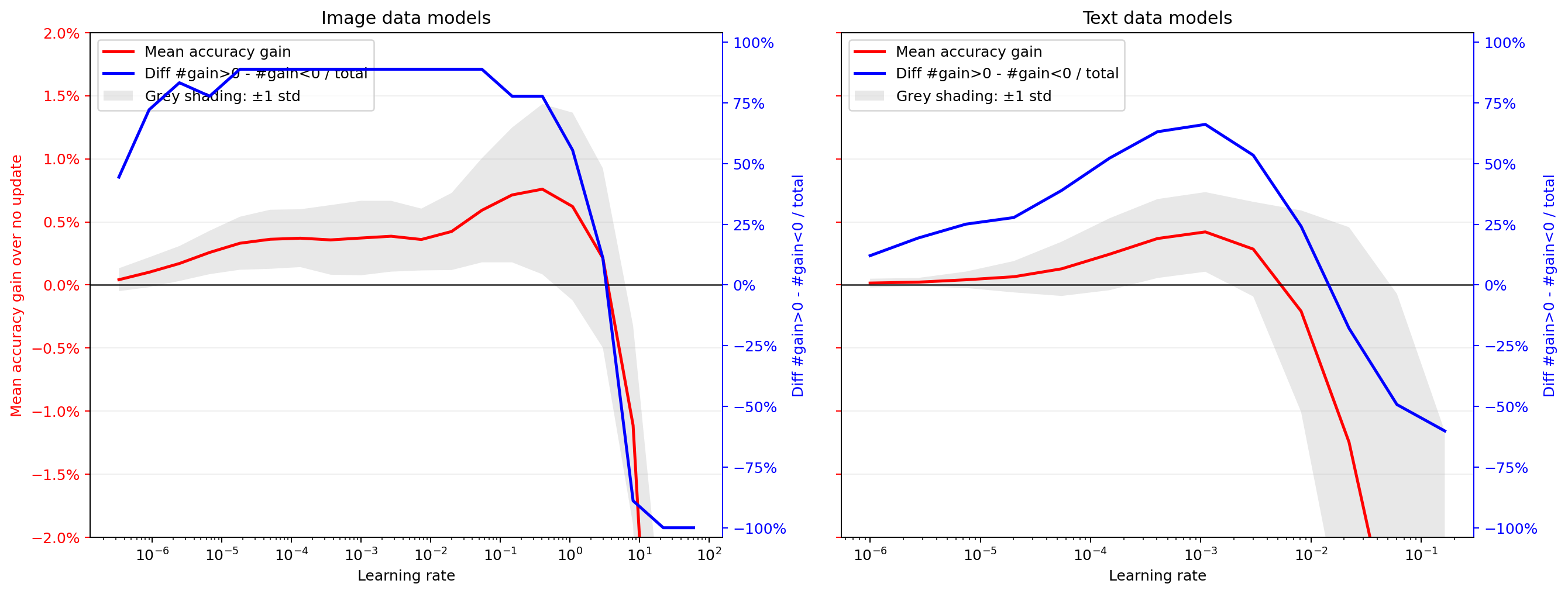}
\vspace{-8pt}
\caption{Average $\Delta_{Acc}$ for different learning rates}
\label{fig:lrIncTe}
\vspace{-6pt}
\end{figure}

\paragraph{Setup.}
We evaluate our methods on more than 300 model-dataset pairs, i.e., vision and language models using ImageNet and diverse text datasets, respectively (see Appendix). We cover output classes ranging from about $10^3$ to more than $10^6$ and models range from below 0.1B to 27B. We use Algorithm \ref{alg:focusopt} mostly with the stated defaults. However, we defined the uncertainty threshold $d_{1,2}$ implicitly. That is, we optimized a fixed percentile of $d_{1,2,\%}$ of all test samples, i.e., the 10\% most uncertain text data meaning 30k samples per dataset (15\% of images meaning 9k samples). This ensures that the  number of uncertain samples is the same across models and datasets. The chosen percentage signifies a trade-off between computational effectiveness and overall accuracy gains as discussed in Experiment 1.   
We optimize using learning rates (lr) $1\mathrm{e}{-6}\cdot 2^i$  for $i\in[0,12]$ for text and pick the best one using a 50:50 validation/test split on a per model basis (for images we use $1/3\mathrm{e}{-6}$ for $i\in[0,20]$). These ranges are enough so that we essentially see no effect for our method for small lr due to Algorithm \ref{alg:focusopt} as well as degradation for large lr as discussed in Experiment 2 and shown in Figure \ref{fig:lrIncTe}.

We compare the optimized model $f^{our}$ with $our \in \{iFo,doFo,Ent\}$ or input $X$ resulting from our optimization (Algorithm \ref{alg:focusopt}) against the original, unmodified model $f$ or input $X$ serving as baseline. We consider various configurations $O= O_{Img} \cup O_{Text}=\{f,D\}$, meaning pairs of models and datasets $(f,D)$ for images $O_{Img}$ and text $O_{Text}$. The test dataset $D_{te}$ is split into sets of uncertain $D_{te,1,2}$  and certain samples. We report measures for individual configurations $(f,D) \in O$ on the entire held-out set, if no hyperparameter tuning took place, otherwise we use two-fold cross-validation.
We commonly use $\Delta_{Acc}$ which typically measures accuracy between one of our methods and the baseline meaning the unmodified model and sample.

\begin{figure}[t]
\centering
\vspace{-6pt}
\includegraphics[width=\linewidth]{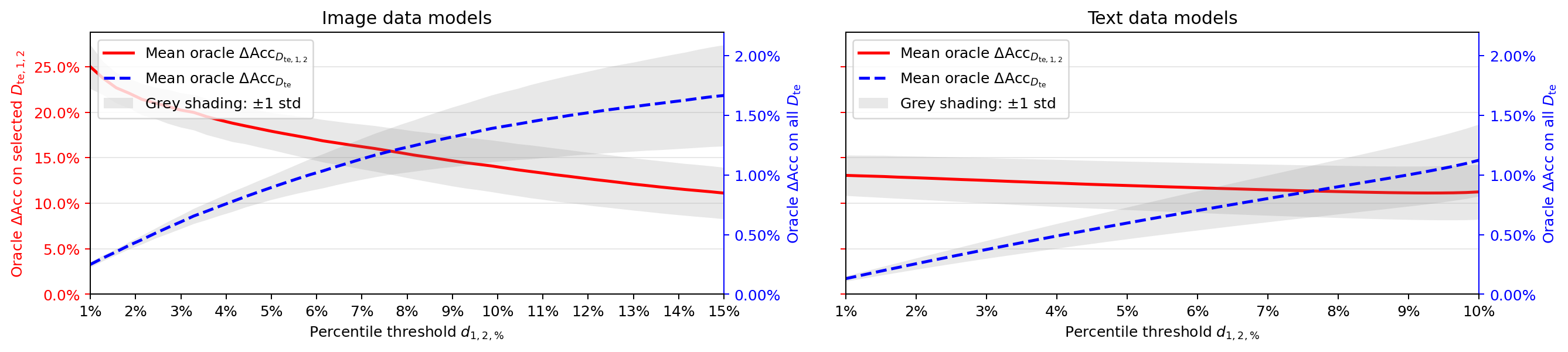}
\vspace{-8pt}
\caption{Larger threshold $d_{1,2,\%}$ implies more samples are uncertain, optimized and fixed so overall accuracy on all samples increases, but computational efficiency and gains on optimized, i.e., uncertain, samples decrease.} 
\label{fig:plotfracMean} 
\vspace{-6pt}
\end{figure}

\textbf{Experiment 1: Oracle gain and Uncertainty fraction:}
We vary the fraction of uncertain samples $d_{1,2,\%}$ and investigate the gain $\Delta_{Acc}$ on all test samples between oracle optimization and no optimization, i.e., an oracle can choose any of the evaluated learning rates for each sample to maximize accuracy. Thus, an oracle prediction is correct if applying just one learning rate yields the correct prediction but it cannot change gradient direction.

\emph{Results} in Figure \ref{fig:plotfracMean} show that oracle gains on uncertain samples ($D_{te,1,2}$, red line) can be as high as 25\% for image data but drop with larger threshold $d_{1,2,\%}$. Larger threshold implies that also more certain samples are considered. Such samples are more likely to be correct, i.e., they cannot be fixed. Thus, the percentage gain decreases on the test data. In turn, this means that efficiency in terms of fixed samples per compute decreases. On all test data $D_{te}$ the oracle gain continues to increase with larger threshold $d_{1,2,\%}$ but the growth gets slower - notably on image data. Also as shown in the next experiment even our best method's gain is more than an order of magnitude below oracle gain.

\textbf{Experiment 2: Our accuracy gains}

Figure \ref{fig:lrIncTe}  shows that for small learning rates, neither significant gains nor losses are observed on average and the number of classifiers that benefit from entropy optimization (positive accuracy gain) is about equal to those that do not. For very large learning rates performance decreases a lot due to optimization and becomes negative. In between we find that both the average gain and the number of models clearly favor optimization.

Table \ref{tab:exp9-all-models} compares the suggested methods for a fixed learning rate per model. Overall entropy performs best gaining on average 0.6\% across all configurations though on about 10\% of all configurations it worsens accuracy. The largest gain of entropy is due to incrementing likely classes (iFo). Just incrementing the top 2 classes ($iFo^2$) yields 2/3 of the entropy's gain, while decrementing less likely classes ($doFo$) leads only to a modest gain of 1/4 of entropy gains. Thus, the idea of removing ``noise'' by decrementing classes ($doFo$) is less beneficial than focusing on likely classes and simply increasing the top-2 classes. Interestingly, there are some models where iFo performs significantly better than entropy (e.g., Gemma), while also performing much worse (e.g., Olmo-2). We investigated several hypotheses related to model architecture and training regime but we could not conclusively factor out the decisive architectural difference. Thus, we opted for a controlled experiment as discussed in the next section. 

\begin{table*}[t]
\centering
\small
\caption{Method comparison}
\label{tab:exp9-all-models}
\resizebox{\textwidth}{!}{%
\begin{tabular}{lcccc}
\toprule
Method & $\Delta_{Acc}$ & \shortstack{$|O|$ with $\Delta_{Acc}$\\pos/neg} & \shortstack{Larger/smaller Acc than Entropy\\pairs} & \shortstack{$\Delta_{Acc}$\\rel. to. Entropy} \\
\midrule
Entropy & $+0.60$ {\scriptsize $(\pm 0.75)$} & 288/37 & -- & -- \\
iFo & $+0.53$ {\scriptsize $(\pm 0.82)$} & 277/48 & 117/206 & $-0.07$ {\scriptsize $(\pm 0.74)$} \\
doFo & $+0.15$ {\scriptsize $(\pm 0.53)$} & 207/117 & 65/260 & $-0.45$ {\scriptsize $(\pm 0.74)$} \\
$iFo^2$ & $+0.41$ {\scriptsize $(\pm 0.71)$} & 261/61 & 99/225 & $-0.19$ {\scriptsize $(\pm 0.67)$} \\
\bottomrule
\end{tabular}
}
\end{table*}

{\textbf{Fixed learning rate (lr) vs gradient-ray ensembling  } \emph{Experiment 2:}
We computed accuracy using our proposed method \emph{gradient-ray ensembling}, i.e.,  summing probabilities across all learning rates and picking the most likely class (see Algorithm \ref{alg:focusopt} and Figure \ref{fig:lat}). For computational reasons, we summed probabilities for the initial top-16 classes and aggregated their probabilities. As shown in the Appendix, this yields much better outcomes than using just the top-4.\footnote{We expect using more classes to behave even better.}. For entropy we get $\Delta_{Acc}=0.75 (\pm 0.80)$ which is a statistically significant (p<0.001) increase compared to using a fixed lr. If we take only a subset of all evaluated learning rates to maximize accuracy (on a validation dataset) and then evaluate on the test data we get $\Delta_{Acc}=0.84 (\pm 0.79)$. More precisely, we started from all learning rates and performed leave-one-lr-out until the gain stopped improving. Gains for methods other than entropy are comparable (see Appendix).

\begin{figure}[ht]
    \centering

    \begin{subfigure}{0.48\linewidth}
    \centering
    \includegraphics[width=\linewidth]{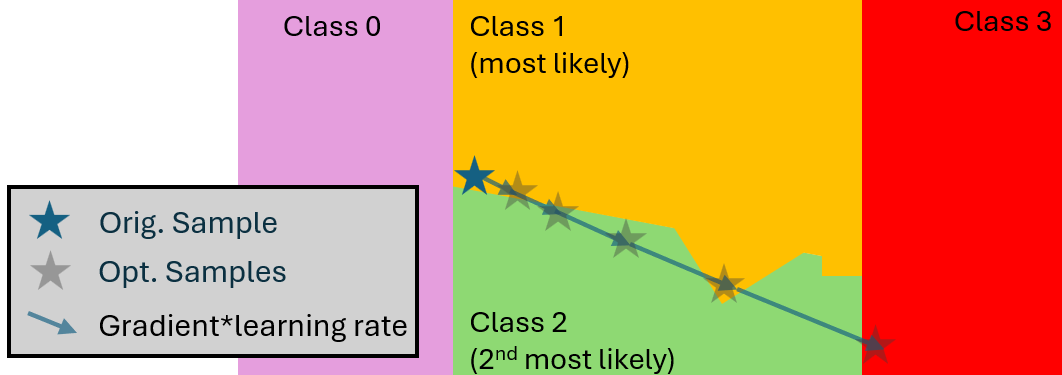}
    \caption{Movement of an uncertain sample when optimized towards few classes (e.g., class 1 and 2) with varying learning rates. It stays close to the decision boundary. Increasing common features tends to push it to the correct class.}
    \label{fig:lat}
    \end{subfigure}
    \hfill
    \begin{subfigure}{0.48\linewidth}
        \centering
        \includegraphics[width=0.65\linewidth]{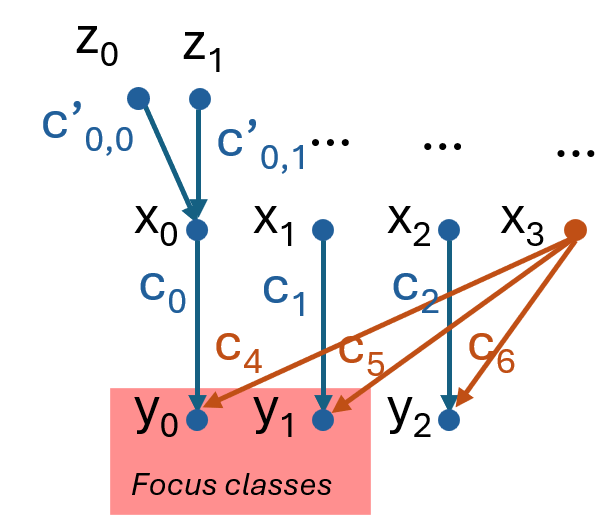}
        \caption{Theoretical model: 3 output classes $y_i$ and one shared feature $x_3$ among all $y_i$}
        \label{fig:theoMod}
    \end{subfigure}
    \caption{Illustrations of input movement and toy model.}
    \label{fig:latAll}
\end{figure}


\section{Understanding Incrementing Classes} \label{sec:th}

We aim to better understand why and how incrementing the logits of the top-2 classes for uncertain samples leads to accuracy gains. We restricted our analysis to the case where the correct class is among the top-2. We performed a controlled experiment that covers a number of basic architectures. We also discuss a more narrow theoretical scenario when accuracy gains can happen.

\subsection{Controlled experiment}
We aim to better understand data and network conditions, in particular architecture effects when maximizing the top-2 classes using raw logits, i.e., we optimize inputs $\frac{\partial {(f_{m(1)}(X)+f_{m(2)}(X))}}{\partial X}$ (and also network parameters $\frac{\partial {(f_{m(1)}(X)+f_{m(2)}(X))}}{\partial \Theta}$). 
We train multiple small networks (below 200k parameters) covering fully connected networks, networks with different normalization layers (none, batchnorm, and layernorm) and w/o residual connections, and a transformer. We focused on character recognition based on EMNIST with 16 classes (0-9,A-F) and trained on roughly 1\% of all data. This ensures that networks perform well ($>90\%$ on test data) but there is still a large number of highly uncertain samples. 

Our analysis reveals that the essential parameter predicting gains of incrementing top-2 classes is the uncertainty threshold $d_{1,2,\%}$. When focusing on highly uncertain samples, i.e., $d_{1,2,\%}=0.02$, we consistently observe accuracy gains across all architectures when optimizing inputs $X$ as well as all network parameters $\Theta$. When focusing on samples that are in the 10\% to 12\% uncertainty percentile accuracy gains vanish when incrementing just raw unweighed logits. Using $iFo$ that employs weighing of classes rather than just using raw logits helps for more certain samples. That is, weighing by probability $p_c$ and logit distance to the mean $\overline(f)(X)-f_c(X)$ reduces the risk that a less likely (wrong) class overtakes a more likely (correct) class, but it also reduces the number of samples that are fixed through optimization. 

For highly uncertain samples $d_{1,2,\%}=0.02$, we also observe that the true class gradient magnitudes dominate those of wrong classes independent of their predicted rank. Assume the correct class is predicted wrongly, i.e., as second most likely. In this case, in more than 50\% of all cases its gradient magnitude is still larger than that of the most likely class. The same holds if the true class is predicted correctly: Its gradient magnitude is largest among the top-2.

\subsection{A possible theoretical rationale}
We investigate unweighed $iFo^2$, i.e., optimizing raw logits without probability weights, denoted as $iFo^{2}_{noPr}$. 

We first prove for a simple model that $iFo^{2}_{noPr}$ makes predictions rely more on shared, non-contrastive features. The example should be easy to follow for a reader, while being rigorous. Proofs and a generalized theory can be found in the Appendix.  A feature $x_i$ is shared between classes $y_0$ and $y_1$ if for a sample $X$ changing the feature's value changes the logits of both classes. A non-contrastive feature for a set of classes $C$ is a shared feature that either increases the logits of all classes $C$ or decreases all of them. We show that $iFo^{2}_{noPr}$ boosts shared features and investigate when it improves predictions. We show that $iFo^{2}_{noPr}$ amplifies the noise-free part of the margin between the two likely classes while leaving the noise unchanged, i.e., it increases the signal-to-noise ratio of the prediction. The change in margin is network dependent, while for a shared, non-contrastive feature there is a wide range of network parameters where $iFo^{2}_{noPr}$ helps. But for shared, contrastive features $iFo^{2}_{noPr}$ can be hurtful.

\paragraph{Model:} We assume three output classes $C=\{y_0,y_1,y_2\}$ and focus classes $F=\{y_0,y_1\}$. We have four input features $X=\{x_0,x_1,x_2,x_3\}$. For an illustration of our model, see Figure \ref{fig:theoMod}.
The input features $x_j$ can be seen as originating from a prior layer $l$ with parameters $c'$ processing activations $z$, i.e.,
$
x_i=l(z|c')=\sum_{j} c'_{j,i}z_j,
$
where $z_j$ are activations for the current sample and $c'_{j,i}$ are trainable parameters.  This corresponds to a pre-activation representation, or equivalently to a post-nonlinearity activation in a locally linear region (e.g., an active ReLU unit). We further assume some form of (layer) normalization, i.e.,
$
\sum_j z_j^2 = 1.
$
To avoid a large number of symmetric case distinctions, we additionally consider a symmetric sample with
$
x_i=1 \quad \text{for all } i.
$
Outputs $y_i:=f_i(X)$ are computed as: 
\begin{align}
    y_0 = c_0 x_0 + c_4 x_3 \phantom{abc}  y_1 = c_1 x_1 + c_5 x_3 \phantom{abc} y_2 = c_2 x_2 + c_6 x_3
\end{align}

We apply the intuitive notion that a feature value cannot be negative, i.e., $x_i\geq 0$, which happens, for example, after layer activations pass through a non-linearity like $ReLU$. We also assume that $c_i\neq 0$ to avoid edge cases that can easily be deduced from our analysis.
The model implicitly expresses that the presence of input features $\{x_0,x_1,x_2\}$ is only indicative of class $i$, i.e., a change of $x_i$ alters only $y_i$ for $i\in\{0,1,2\}$. In turn, $c_0>0$, $c_1>0$, and $c_2>0$. The shared feature $x_3$ impacts all outputs $y_i$. It might also be contrastive, i.e., it can be that two parameters from $\{c_4, c_5,c_6\}$ have opposing signs.
Depending on the sign of $c_4$, $c_5$ and $c_6$ the presence of the feature ($x_3>0$) increases or decreases $y_i$. As we model dependence among classes only through the shared feature $x_3$, we assume that parameters $c'_{\cdot,i}$ are feature-specific, so that updating the parameters producing $x_i$ does not directly affect $x_m$ for $m\neq i$. This also facilitates our analysis.

Thus, we do not show any preference for a shared or non-shared feature and its upstream features $z_i$. We discuss deviating from this assumption after the proofs as they clarify the core behavior.  We keep the head coefficients $c_i$ fixed and update only the upstream parameters $c'_{j,i}$ (as our implementation tunes embeddings/inputs). We use the loss from our method $iFo^{2}_{noPr}$ $L^{iFo^{2}_{noPr}}=-(y_0+y_1)$ without weighting and the loss from optimizing just a single class $L_{y_i,-}=-y_i$. Since $y_i$ is linear in $x$ and $x$ is linear in $c'$, one gradient step changes the logits exactly. We write the learning rate as $\eta$. The theorems are stated for $\eta=1$ and the proofs give the dependence on $\eta$.  
For an illustration of our model, see Figure \ref{fig:theoMod}.

\begin{theorem}\label{thm:one}
For a non-contrastive shared feature, the update to the shared feature is larger in magnitude for unweighed $iFo^{2}_{noPr}$ than for the most likely class: $|c_4+c_5|>\max(|c_4|,|c_5|)$. For a contrastive feature and most likely class $y_0$, the $iFo^{2}_{noPr}$ update is smaller in magnitude than for the most likely class iff $|c_5|<2|c_4|$. For a non-shared feature $x_i$ of the most likely class $y_i$ updates to $x_i$ are identical.
\end{theorem}

Next, we investigate when $iFo^{2}_{noPr}$ improves predictions (and when it does not).
Assume all parameters originating from training are perfect, so that without noise the predictions are always correct. This can easily be achieved using sufficient training. Furthermore, if performance on training data is poor, there are chances that further training might improve also test-time performance - in particular, as overfitting is less of a concern in deep learning \cite{bel19,nak20}. At test time, we assume the presence of noise, e.g., to account for differences between training and test data distribution.

As shown in the Preliminaries in the Appendix, the $iFo^{2}_{noPr}$ update $\Delta x_i$ does not depend on the values $x_j$ but only on parameter values $c_i$, so additive noise on any feature passes through the update unchanged; in particular, we do not need to assume that the shared feature $x_3$ is noise-free. For concreteness, assume only $x_0$ is noisy (the cases of noise on $x_1$ or $x_3$ are analogous, see Remark \ref{rem:noise}), i.e.
\[
\hat{x}_0 = x_0 + R_0, \qquad R_0 \sim \mathrm{Uniform}[-1,1].
\]
Also, coefficients $c_0=c_2=c_5=1$. The coefficient $c_4$ for the shared feature is $c_4=1+\delta_4$ and for the non-shared feature $x_1$ we have $c_1=1+\delta_1$.  
Then
\[
\hat{y}_0 = y_0 + c_0 R_0 = y_0 + R_0.
\]
As we do not constrain $\delta_1,\delta_4$ to be either positive or negative, this setup covers many cases.  $\delta_1$ and $\delta_4$ are assumed to arise from the training data. For example, consider binary features and a deterministic frequency-based linear classifier, where the coefficient of a feature is a monotone function of its frequency in the corresponding class (e.g., the log-odds under Bernoulli Naive Bayes). If the shared feature $x_3$ occurs slightly more often for class $y_0$ than the class-specific feature $x_1$ occurs for $y_1$, training yields $c_4>c_1$, and hence $\delta_4>\delta_1$ (and vice versa). Thus, a margin $m^*=\delta_4-\delta_1$ (expressed later in Theorem \ref{thm:two}) can be viewed as a small asymmetry learned from the training data, which $iFo^{2}_{noPr}$ subsequently amplifies without amplifying the additive test noise.

It is also motivated by the fact that these assumptions lead to similar  outputs $y_i$ and thus high uncertainty if $|\delta_4|$ and $|\delta_1|$ are small. In the most extreme case, when $\delta_4=\delta_1=0$ we get maximal uncertainty without noise $R_0=0$,  meaning $y_0=y_1$. To avoid this edge case, we assume $\delta_1\neq \delta_4$. 
Note, that a key quantity in our analysis is $c_4+c_5=(1+\delta_4)+1$. We call a feature positively shared if $\delta_4>-1$ so that $c_4>0$ and $c_5>0$ and, thus, $|c_4+c_5| > \max(|c_4|,|c_5|)$. 
Let us focus on the situation, where the correct class is within the top-k classes and thus only analyze the outputs $y_0,y_1$. This is the most interesting case, as otherwise the prediction on the original classifier is already wrong (and thus changing the prediction in any way cannot cause a wrong prediction as it is already wrong, i.e., $iFo^{2}_{noPr}$ might only help in terms of the 0/1 loss of this sample, although it may still reduce the margin of the correct class). For this, class $y_2$ must neither be among the two most likely classes before the step nor overtake after it, i.e., $y_2<\min(y_0,y_1)$ and $y_2+\Delta y_2<\max(y_0+\Delta y_0,\,y_1+\Delta y_1)$. Since $\Delta y_2=\eta c_6(c_4+c_5)=\eta c_6(2+\delta_4)$, while $\Delta y_0=\eta(3+3\delta_4+\delta_4^2)>0$ and $\Delta y_1=\eta\big(2+\delta_4+(1+\delta_1)^2\big)$ (see the proof of Theorem \ref{thm:two}), both conditions hold for all $R_0\in[-1,1]$ if, e.g., $c_6<\min\{0,\delta_4,1+\delta_1\}$ and $\delta_4>-2$.
\begin{theorem}[$iFo^{2}_{noPr}$ amplifies the noise-free margin]\label{thm:two}
Let $m^\ast:=\delta_4-\delta_1\neq0$ be the noise-free gap $y_0-y_1$, so that the correct class is $y_0$ if $m^\ast>0$ and $y_1$ if $m^\ast<0$, and let
\[
a:=1+\eta(2+\delta_4+\delta_1).
\]
One $iFo^{2}_{noPr}$ step with learning rate $\eta$ turns the gap $y_0-y_1=R_0+m^\ast$ into
\[
(y_0+\Delta y_0)-(y_1+\Delta y_1)=R_0+a\,m^\ast .
\]
The noise is unchanged and the noise-free gap is scaled by $a$. Hence $iFo^{2}_{noPr}$ improves the margin of the correct class iff $a>1$, i.e., iff $\delta_4+\delta_1>-2$; this holds in particular whenever $x_3$ is positively shared ($\delta_4>-1$) and $c_1>0$ ($\delta_1>-1$). If $\delta_4+\delta_1<-2$, the margin shrinks, and for $a<0$ the noise-free ordering is reversed.
\end{theorem}
\begin{corollary}[Accuracy]\label{cor:acc}
For $R_0\sim\mathrm{Uniform}[-1,1]$, the probability of deciding correctly between $y_0$ and $y_1$ is $\tfrac12\min\{2,\max\{0,\,1+a|m^\ast|\}\}$; before the step ($a=1$) this is $\tfrac12(1+|m^\ast|)$ for $|m^\ast|\le1$, and it is non-decreasing in $a$.
\end{corollary}
In terms of the original coefficients ($c_0=c_5=1$): the noise-free gap is $c_4-c_1$ and $iFo^{2}_{noPr}$ adds $\eta(c_4^2-c_1^2)=\eta(c_4-c_1)(c_4+c_1)$, so the condition $\delta_4+\delta_1>-2$ is simply $c_4+c_1>0$. Thus the non-shared coefficient $c_1$ may be arbitrary as long as it is positive, which is no restriction for a feature with $x_1>0$ that is indicative of class $y_1$. To improve accuracy the toy experiment requires that the shared feature is more prevalent in the correct class. We conjecture: Accuracy improves by relying more on shared features as shared features are more prevalent in the correct class.

\section{Related Work} \label{sec:relat}

To our knowledge, almost all prior works (see Table \ref{tab:RelWork} in Appendix) have extra needs beyond the test sample. They require  additional tasks \citep{liu21tt,sun20te} or data (potentially, self-generated using augmentations (Cotta, Memo, DeYO \citep{wa22cotta,zha22memo,lee24deyo}) or in-domain data \citep{liu21tt} or in-domain samples\citep{hu25reca}). We intentionally focus on just a single test sample and refrain from such extra needs as they limit practical applicability, require effort, and often pose risks.
Many studies on domain adaptation (see Table \ref{tab:RelWork} in Appendix for a summary) and others \citep{kim24te,oso24,kar24}) assume that test-time data originates from a domain different from that of the training data, whereas we focus on in-domain samples. Closest to our work in terms of problem setup is Memo\cite{zha22memo} as it also stresses robustness more broadly and it evaluates on in-domain samples. But its gains for in-domain samples rely on augmentations. 
Methodologically, many TTA methods are similar to one another and to ours (see Table \ref{tab:metcomp} in Appendix). We differ by parameters optimized (we optimize inputs or their embeddings), loss functions (we dissect entropy) and, more interestingly, we propose a form of self-ensembling (gradient-ray ensembling). Some overlaps such as sample splitting appear coincidental as they serve different purposes or lead to opposing decisions. SAR optimizes only certain samples (measured by entropy), whereas we optimize the opposite, i.e., uncertain samples. PASLE, among other works in TTA, investigated focusing on more than one target class, e.g., hard labels for likely classes \citep{hu25pals} and entropy minimization \citep{wan20tent,zha22memo,hu25reca}.
Commonly, test-time tuning relies on nearest neighbors and conducts a form of local learning by adjusting the model to the chosen data such as TSD+MLSC \citep{wan23fea} and others \citep{sun24le,bot92,har23te,hub24}. \citep{mum21te} aims at accounting for distribution shifts using the soft likelihood ratio (SLR) loss. Conceptually, iFo and doFo are multi-class optimizations, where we consider all focus (and out-of-focus) classes as target classes.} Our technique is reflective as it often does not yield a decision based on a single forward pass \citep{zho24,sch25gen,mad24,sch24ref,sel17}.
When tuning embedding vectors, our technique can also be seen through the lens of moving the embedding vectors in a direction. This interpretation is well-known for word vectors with some evidence \citep{miko13} for more abstract function vectors in LLMs \citep{todd24}.
Further related work discussing ``Test-time optimization of prompts and latent representations'' and ``{Optimizing inputs instead of model parameters.}'' is in the Appendix.

\section{Limitations, Discussion and Conclusions} \label{sec:discu}
This work set out to answer the question \emph{Does focusing on likely classes of a single, in-domain sample improve model predictions?} Applying entropy optimization yields ``yes'' on average, but gains are model, data, and uncertainty threshold dependent. Our study also hints that applying any test time adaptation technique on just say 3 networks and datasets has limited implications on its generality, as in our case it would be easy to pick a few datasets and models that make any of the methods iFo, $iFo^2$, doFo, and Entropy look much more favorable than they might appear across a large number of datasets and models. Still, despite using a large number of datasets and models the same criticism also applies to our study, i.e., applying even more datasets and more models might change outcomes. For example, while we included model families with different sizes, e.g., "Qwen/Qwen3.5-2B-Base" and its 9B version, and found no clear differences, very large models might alter our findings.
Surprisingly simply increasing the two most likely classes captures most of the benefit of entropy minimization suggesting that gains are less due to reducing ``unorderedness'' as the term implies but rather due to lifting the right candidate classes. In particular, our controlled experiment showed that the gradient magnitude of the correct class tends to be larger. Our theory showed a set of conditions, when increasing the two most likely classes improves predictions, but other scenarios are likely plausible as well. 

Furthermore by aggregating probabilities of classes for different learning rates (gradient-ray ensembling) gains can be further increased, which also is worth studying in more depth.  We used softmax probabilities, for which calibration is a known issue\citep{guo17} though less for pre-trained LLMs \citep{xie24cal}. Applying calibration techniques is future work.

\subsection*{AI use statement}

Yes, to aid or polish writing.
AI (ChatGPT 5+6 and Claude Opus+Claude) was used for spelling and grammar checks.

Yes, for retrieval and discovery (e.g., finding related work). Most references were not found by AI, but a good number originate from AI. Related work was manually written except for 3 sentences in the appendix.

Yes, for research ideation or execution: A significant amount of code was AI-generated. The original algorithm was manually done, but later optimization and evaluation were mostly AI-generated, in particular the controlled experiment.

Yes, for proving mathematical claims. The toy example was mostly done manually, but refined by AI. The general theory in the Appendix included in the supplement submission was mostly AI-generated.

\bibliographystyle{iclr2027_conference}
\bibliography{refs}

\appendix

\begin{figure}[!h]
\centering
\includegraphics[width=1.05\linewidth]{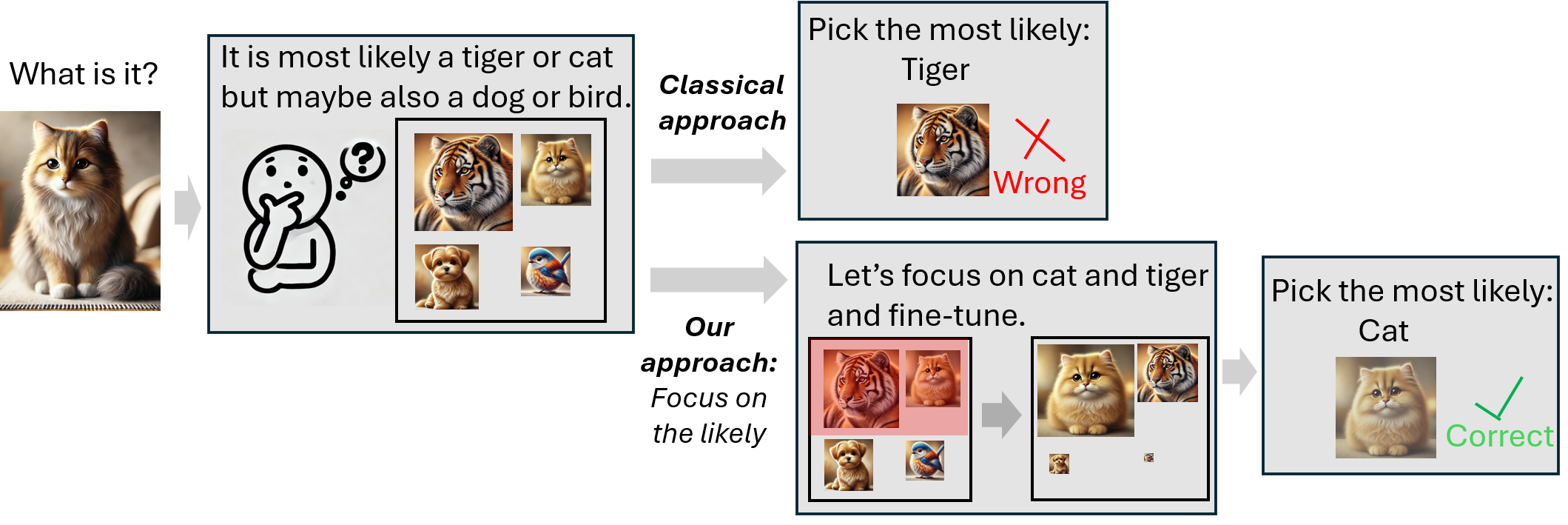}
\vspace{-6pt}
\caption{Overview: Fine-tune based on likely initial classes.}
\vspace{-6pt}
\label{fig:overhigh}
\end{figure}

\begin{table}[h!]
\centering
{
\scriptsize
\setlength{\tabcolsep}{3pt}
\begin{tabular}{l c c c c c c c}
& \multicolumn{2}{c}{\textbf{In-Domain}} & \multicolumn{2}{c}{\textbf{1 Test Sample}}  &\textbf{No extras} &  \\  
\cmidrule(lr){2-3} \cmidrule(lr){4-5}
\textbf{Paper}  & Key  & Eval & Per batch /  & Model  & \\
&  topic & &  Eval & Reset & \\

\midrule
Cotta, TTT++,TSD+MLSC & $\times$ & $\times$ & $\times$/$\times$ & $\times$ & $\times$ \\
TTT, ReCAP, DeYO        & $\times$ & $\times$ & \checkmark/\checkmark  & $\times$ & $\times$ \\ 
Memo                    & (\checkmark) & (\checkmark) &  \checkmark/$\times$ & $\times$ & $\times$ \\
Tent, PASLE             & $\times$ & \checkmark & \checkmark/$\times$ & $\times$ & \checkmark  \\
SAR                     & $\times$ & $\times$ & \checkmark/\checkmark & $\times$ & \checkmark  \\
\textbf{Ours}           & \checkmark & \checkmark & \checkmark/\checkmark & \checkmark & \checkmark \\
\end{tabular}
}
\caption{{Works on similar problems discussed in related work.}}

\label{tab:RelWork}
\vspace{-8pt}
\end{table}

\begin{table*}[h]
\centering
\scriptsize
\vspace{-6pt}
\caption{{Methodological comparison with adaptation methods.}} \label{tab:metcomp}
\vspace{-6pt}
{\begin{tabular}{lccccc}
\textbf{Aspect} & PASLE & Tent  &ReCAP& SAR & \textbf{Ours} \\
\midrule 
Loss Function  & Custom & Entropy &  (localized) Entropy  & (reliable,flat) Entropy  & Entropy, iFo, doFo, $iFo^2$ \\
Parameters optimized & All & Norm. Layers & Norm. Layers & Norm. Layers & Inputs (or their embeddings)\\
Pseudo-labels  & Hard, custom loss & - & - & - & Hard, soft weighting  \\
Sample splitting & \checkmark&  $\times$& $\times$  & \checkmark  & \checkmark \\
Optimize only subset & $\times$ &  $\times$& $\times$  & \checkmark(certain)  & \checkmark(uncertain) \\
No extra needs & \checkmark & $\times$(augment) & $\times$ (in-domain data) & \checkmark & \checkmark \\
Self-ensembling & $\times$ & $\times$(augment) & $\times$ $\times$ & $\times$ & \checkmark(gradient-ray) \\
\end{tabular}}
\vspace{-6pt}
\end{table*}

\section{Theory for toy example}
\paragraph{Preliminaries: }
To study changes in the features, we consider one gradient step on the upstream parameters $c'_{j,i}$. Since
$
x_i=\sum_j c'_{j,i} z_j,
$
we have
\[
\frac{\partial L}{\partial c'_{j,i}}= \frac{\partial L}{\partial x_i}\frac{\partial x_i}{\partial c'_{j,i}}=\frac{\partial L}{\partial x_i} z_j.
\]
Hence, for learning rate $\eta$,
\begin{footnotesize}
\begin{align}
\Delta x_i
=
\sum_j \Delta c'_{j,i} z_j
=
-\eta \sum_j \frac{\partial L}{\partial c'_{j,i}} z_j
=
-\eta\frac{\partial L}{\partial x_i}\sum_j z_j^2
=
-\eta\frac{\partial L}{\partial x_i}.
\end{align}
\end{footnotesize}
Note that $\partial L/\partial x_i$ depends only on the coefficients $c_i$ and not on the feature values; hence $\Delta x_i$ is the same for noisy and noise-free features.
The feature-level derivatives are therefore
\begin{footnotesize}
\begin{align}
\frac{\partial L_{y_j,-}}{\partial x_i}
&=
\begin{cases}
-c_j, & i=j,\\
-c_{4+j}, & i=3,\\
0, & \text{otherwise},
\end{cases}
\\
\frac{\partial L^{iFo^{2}_{noPr}}}{\partial x_i}
&=
\begin{cases}
-c_0, & i=0,\\
-c_1, & i=1,\\
-(c_4+c_5), & i=3,\\
0, & \text{otherwise}.
\end{cases}
\end{align}
\end{footnotesize}
Thus, for $\eta=1$, the induced feature changes are
\begin{align}
\Delta x_i^{\,y_j}
&=
\begin{cases}
c_j, & i=j,\\
c_{4+j}, & i=3,\\
0, & \text{otherwise},
\end{cases}
\\[0.5em]
\Delta x_i^{\,iFo^{2}_{noPr}}
&=
\begin{cases}
c_0, & i=0,\\
c_1, & i=1,\\
c_4+c_5, & i=3,\\
0, & \text{otherwise}.
\end{cases}
\end{align}
\begin{theorem}\label{thm:one}
For a non-contrastive shared feature, the update to the shared feature is larger in magnitude for $iFo^{2}_{noPr}$ than for the most likely class: $|c_4+c_5|>\max(|c_4|,|c_5|)$. For a contrastive feature and most likely class $y_0$, the $iFo^{2}_{noPr}$ update is smaller in magnitude than for the most likely class iff $|c_5|<2|c_4|$. For a non-shared feature $x_i$ of the most likely class $y_i$ updates to $x_i$ are identical.

\end{theorem}
\begin{proof}
By definition, a non-contrastive feature $x_3$ either increases or decreases both logits, thus $\mathrm{sgn}(c_4)=\mathrm{sgn}(c_5)$.
For the shared feature $x_3$ for $iFo^{2}_{noPr}$ we have that its change is $\Delta x_3=c_4 + c_5$ and for the most likely class it is either $c_4$ or $c_5$ depending on whether $y_0$ or $y_1$ is the most likely class. If a feature is non-contrastive then we have $\mathrm{sgn}(c_4)=\mathrm{sgn}(c_5)$ and thus $|c_4+c_5|=|c_4|+|c_5| > \max(|c_4|,|c_5|)$, i.e., the update is larger for non-contrastive shared features for $iFo^{2}_{noPr}$ ($k=2$) than when updating the most likely class $k=1$ (which can be either $y_0$ or $y_1$).
If $\mathrm{sgn}(c_4)\neq \mathrm{sgn}(c_5)$, then $|c_4+c_5| = \big||c_4|-|c_5|\big|$, which is smaller than $|c_4|$ iff $|c_5|<2|c_4|$, and for $|c_5|>2|c_4|$ we have $|c_4+c_5|=|c_5|-|c_4|>|c_4|$ with $\mathrm{sgn}(c_4+c_5)=\mathrm{sgn}(c_5)$.
For non-shared features $x_0,x_1$ the update is $\Delta x_j=c_j$ for $j \in \{0,1\}$ under both losses.
\end{proof}

Next, we investigate when $iFo^{2}_{noPr}$ improves predictions (and when it does not).
Assume all parameters originating from training are perfect, so that without noise the predictions are always correct. This can easily be achieved using sufficient training. Furthermore, if performance on training data is poor, there are chances that further training might improve also test-time performance - in particular, as overfitting is less of a concern in deep learning (\cite{bel19,nak20}). At test time, we assume the presence of noise, e.g., to account for differences between training and test data distribution.

As shown in the Preliminaries, the $iFo^{2}_{noPr}$ update $\Delta x_i$ does not depend on the feature values, so additive noise on any feature passes through the update unchanged; in particular, we do not need to assume that the shared feature $x_3$ is noise-free. For concreteness, assume only $x_0$ is noisy (the cases of noise on $x_1$ or $x_3$ are analogous, see Remark \ref{rem:noise}), i.e.
\[
\hat{x}_0 = x_0 + R_0, \qquad R_0 \sim \mathrm{Uniform}[-1,1].
\]
Also, coefficients $c_0=c_2=c_5=1$. The coefficient $c_4$ for the shared feature is $c_4=1+\delta_4$ and for the non-shared feature $x_1$ we have $c_1=1+\delta_1$.
Then
\[
\hat{y}_0 = y_0 + c_0 R_0 = y_0 + R_0.
\]
As we do not constrain $\delta_1,\delta_4$ to be either positive or negative, this setup covers many cases. 
It is also motivated by the fact that these assumptions lead to similar  outputs $y_i$ and thus high uncertainty if $|\delta_4|$ and $|\delta_1|$ are small. In the most extreme case, when $\delta_4=\delta_1=0$ we get maximal uncertainty without noise $R_0=0$,  meaning $y_0=y_1$. To avoid this edge case, we assume $\delta_1\neq \delta_4$. 
Note, that a key quantity in our analysis is $c_4+c_5=(1+\delta_4)+1$. We call a feature positively shared if $\delta_4>-1$ so that $c_4>0$ and $c_5>0$ and, thus, $|c_4+c_5| > \max(|c_4|,|c_5|)$. 
Let us focus on the situation, where the correct class is within the top-k classes and thus only analyze the outputs $y_0,y_1$. This is the most interesting case, as otherwise the prediction on the original classifier is already wrong (and thus changing the prediction in any way cannot cause a wrong prediction as it is already wrong, i.e., $iFo^{2}_{noPr}$ might only help in terms of the 0/1 loss of this sample, although it may still reduce the margin of the correct class). For this, class $y_2$ must neither be among the two most likely classes before the step nor overtake after it, i.e., $y_2<\min(y_0,y_1)$ and $y_2+\Delta y_2<\max(y_0+\Delta y_0,\,y_1+\Delta y_1)$. Since $\Delta y_2=\eta c_6(c_4+c_5)=\eta c_6(2+\delta_4)$, while $\Delta y_0=\eta(3+3\delta_4+\delta_4^2)>0$ and $\Delta y_1=\eta\big(2+\delta_4+(1+\delta_1)^2\big)$ (see the proof of Theorem \ref{thm:two}), both conditions hold for all $R_0\in[-1,1]$ if, e.g., $c_6<\min\{0,\delta_4,1+\delta_1\}$ and $\delta_4>-2$.
\begin{theorem}[$iFo^{2}_{noPr}$ amplifies the noise-free margin]\label{thm:two}
Let $m^\ast:=\delta_4-\delta_1\neq0$ be the noise-free gap $y_0-y_1$, so that the correct class is $y_0$ if $m^\ast>0$ and $y_1$ if $m^\ast<0$, and let
\[
a:=1+\eta(2+\delta_4+\delta_1).
\]
One $iFo^{2}_{noPr}$ step with learning rate $\eta$ turns the gap $y_0-y_1=R_0+m^\ast$ into
\[
(y_0+\Delta y_0)-(y_1+\Delta y_1)=R_0+a\,m^\ast .
\]
The noise is unchanged and the noise-free gap is scaled by $a$. Hence $iFo^{2}_{noPr}$ improves the margin of the correct class iff $a>1$, i.e., iff $\delta_4+\delta_1>-2$; this holds in particular whenever $x_3$ is positively shared ($\delta_4>-1$) and $c_1>0$ ($\delta_1>-1$). If $\delta_4+\delta_1<-2$, the margin shrinks, and for $a<0$ the noise-free ordering is reversed.
\end{theorem}
\begin{corollary}[Accuracy]\label{cor:acc}
For $R_0\sim\mathrm{Uniform}[-1,1]$, the probability of deciding correctly between $y_0$ and $y_1$ is $\tfrac12\min\{2,\max\{0,\,1+a|m^\ast|\}\}$; before the step ($a=1$) this is $\tfrac12(1+|m^\ast|)$ for $|m^\ast|\le1$, and it is non-decreasing in $a$.
\end{corollary}
In terms of the original coefficients ($c_0=c_5=1$): the noise-free gap is $c_4-c_1$ and $iFo^{2}_{noPr}$ adds $\eta(c_4^2-c_1^2)=\eta(c_4-c_1)(c_4+c_1)$, so the condition $\delta_4+\delta_1>-2$ is simply $c_4+c_1>0$. Thus the non-shared coefficient $c_1$ may be arbitrary as long as it is positive, which is no restriction for a feature with $x_1>0$ that is indicative of class $y_1$.
\begin{proof}
By the Preliminaries, $\Delta x_0=\eta c_0$, $\Delta x_1=\eta c_1$, $\Delta x_3=\eta(c_4+c_5)$, independently of $R_0$. Since the head is fixed, $\Delta y_0=c_0\Delta x_0+c_4\Delta x_3=\eta\big(c_0^2+c_4(c_4+c_5)\big)$ and $\Delta y_1=\eta\big(c_1^2+c_5(c_4+c_5)\big)$, hence
\[
\Delta y_0-\Delta y_1=\eta\big[(c_0^2-c_1^2)+(c_4^2-c_5^2)\big]
=\eta\big[\delta_4(2+\delta_4)-\delta_1(2+\delta_1)\big]
=\eta(\delta_4-\delta_1)(2+\delta_4+\delta_1)=(a-1)\,m^\ast .
\]
Adding the original gap $R_0+m^\ast$ gives $R_0+a\,m^\ast$. Since $a$ is increasing in $\delta_4+\delta_1$ with $a=1$ at $\delta_4+\delta_1=-2$ and $a=0$ at $\delta_4+\delta_1=-2-1/\eta$, the remaining claims follow; $\delta_4>-1$ and $\delta_1>-1$ imply $\delta_4+\delta_1>-2$. For the corollary, if $m^\ast>0$ the decision is correct iff $R_0>-a\,m^\ast$, and $\Pr(R_0>-t)=\tfrac12\min\{2,\max\{0,1+t\}\}$ for $R_0\sim\mathrm{Uniform}[-1,1]$; the case $m^\ast<0$ is symmetric.
\end{proof}
Remark 1: For $\delta_4 = \delta_1$ (in particular for perfect symmetry $\delta_4=\delta_1=0$), $iFo^{2}_{noPr}$ has no impact, i.e.,
Before $iFo^{2}_{noPr}$,
\[
y_0 - y_1 = R_0.
\]
After $iFo^{2}_{noPr}$,
\[
(y_0 + \Delta y_0) - (y_1 + \Delta y_1) = R_0 .
\]
Thus, the ground truth assigns equal likelihood to both classes, and the outcome depends only on noise. In this case, $iFo^{2}_{noPr}$ has no effect.
Remark 2: Deviating from a symmetric setup and using $x_3\neq 1$ (with $x_0=x_1=1$) does not change the update, because $\Delta x_i$ does not depend on the feature values: the gap after the step is
\[
R_0+\delta_4x_3-\delta_1+\eta(\delta_4-\delta_1)(2+\delta_4+\delta_1).
\]
The feature value $x_3$ only enters the noise-free gap $\delta_4x_3-\delta_1$. For $\delta_1=0$ the gap is $R_0+\delta_4\big(x_3+\eta(2+\delta_4)\big)$: $iFo^{2}_{noPr}$ improves the margin iff $\delta_4>-2$, independently of $x_3$, and a larger $x_3$ only moves the threshold for reversing the noise-free ordering from $\delta_4>-2-1/\eta$ to $\delta_4>-2-x_3/\eta$. For $\delta_1\neq0$ $iFo^{2}_{noPr}$ improves the margin iff $\mathrm{sgn}(\delta_4x_3-\delta_1)=\mathrm{sgn}\big((\delta_4-\delta_1)(2+\delta_4+\delta_1)\big)$, which a larger $x_3$ can either help or hurt (e.g., $\delta_4=-1.2$, $\delta_1=-0.9$: the condition holds for $x_3=\tfrac12$ but not for $x_3\in\{1,2\}$). Preferring the shared feature in the update (a larger $\sum_jz_j^2$ for the parameters producing $x_3$, say $g>1$) multiplies the shared part of the gap change, giving $\eta\big[-\delta_1(2+\delta_1)+g\,\delta_4(2+\delta_4)\big]$ instead of $\eta(\delta_4-\delta_1)(2+\delta_4+\delta_1)$; for $\delta_1=0$ this changes the size but not the sign condition $\delta_4>-2$.
\begin{remark}[Noise on other features]\label{rem:noise}
If instead $x_1$ or $x_3$ is noisy, $\hat x_1=1+R_1$ or $\hat x_3=1+R_3$, the gap before the step is $R'+\delta_4-\delta_1$ with $R'=-c_1R_1=-(1+\delta_1)R_1$, respectively $R'=(c_4-c_5)R_3=\delta_4R_3$, and Theorem \ref{thm:two} holds with $R_0$ replaced by $R'$ (and Corollary \ref{cor:acc} with the distribution of $R'$), because the update does not depend on the feature values.
\end{remark}
\begin{remark}[Probability-weighted loss]
With the weights $\tilde p_0,\tilde p_1$ of the main paper the change of the gap is
\[
\frac{\eta}{2}\Big[(\tilde p_0+\tilde p_1)(\delta_4-\delta_1)(2+\delta_4+\delta_1)+(\tilde p_0-\tilde p_1)\big(1+(1+\delta_1)^2+\delta_4^2\big)\Big].
\]
For $\tilde p_0=\tilde p_1$ (the uncertain case) Theorem \ref{thm:two} holds with $\eta$ replaced by $\eta\tilde p_0$. For $\tilde p_0\neq\tilde p_1$ the second term has the sign of $\tilde p_0-\tilde p_1$, i.e., of the current noisy gap $R_0+m^\ast$, and favors the currently leading class; it vanishes at a tie.
\end{remark}
\begin{remark}[Updating the head as well]\label{rem:head}
If the head coefficients are updated too ($\Delta c_0=\eta\hat x_0=\eta(1+R_0)$, $\Delta c_1=\eta x_1$, $\Delta c_4=\Delta c_5=\eta x_3$), the head update adds $\eta(\hat x_0^2-x_1^2)=\eta(2R_0+R_0^2)$ to the gap, which is pure noise since $x_0=x_1=1$ without noise, and the products of head and feature updates add $\eta^2(R_0-\delta_1)$. Grouping the exact gap after the step into its noise-free and noise parts,
\[
(y_0+\Delta y_0)-(y_1+\Delta y_1)=\big[m^\ast\big(1+\eta(2+\delta_4+\delta_1)\big)-\eta^2\delta_1\big]+\big[(1+\eta)^2R_0+\eta R_0^2\big].
\]
Compared with Theorem \ref{thm:two}, the noise-free gap is amplified by the same factor (up to $-\eta^2\delta_1$), but the noise is amplified as well: its coefficient grows from $1$ to $(1+\eta)^2$ and it acquires the bias $E[\eta R_0^2]=\eta/3$ in favor of the class with the noisy feature. For $\eta=1$ and $R_0\sim\mathrm{Uniform}[-1,1]$ the standard deviation of the noise grows by the factor $\sqrt{16/3+4/45}/\sqrt{1/3}\approx4.0$ while the noise-free gap grows by $3+\delta_4+\delta_1-\delta_1/m^\ast$, so the signal-to-noise ratio improves only for $\delta_4+\delta_1\gtrsim1$ (for $\delta_1=0$: $\delta_4>1.03$) instead of $\delta_4+\delta_1>-2$. This is why we keep the head fixed.
\end{remark}

\section{Further related work}
\paragraph{Test-time optimization of prompts and latent representations.}
Several works are particularly related in that they keep the pretrained model fixed
and optimize continuous representations at inference time. Test-Time Prompt Tuning
(TPT) learns continuous prompt embeddings from a single unlabeled test example by
entropy minimization \citep{shu22tpt}. Its entropy objective is therefore closely
related to ours, but TPT obtains its self-supervision from confidence-selected
augmentations and optimizes auxiliary prompt tokens rather than the representation
of the observed input itself. In language models, Plug-and-Play Language Models
(PPLM) use gradients from an attribute objective to modify hidden activations during
generation while keeping the LM parameters fixed \citep{dathathri20pplm}.
Hidden-State Optimization (HSO) similarly performs gradient-based inference-time
updates of cached hidden states rather than model parameters
\citep{yoshida21hso}; its gradient, however, is obtained from the likelihood of
observed evaluation tokens rather than solely from the model's current uncertain
prediction. COLD decoding optimizes continuous relaxed token representations by
gradient-based sampling under externally specified energy functions
\citep{qin22cold}. These approaches establish inference-time optimization in
continuous representation space, but use either auxiliary control objectives,
observed tokens, augmentations, or generation constraints.

Recent work has moved this idea toward self-guided latent optimization in LLMs.
Latent Thought Policy Optimization (LTPO) treats intermediate latent thought vectors
as per-instance optimization variables while freezing all model parameters and uses
an intrinsic confidence-based reward computed from the LLM's own output distribution
\citep{ye26ltpo}. This is close to our single-sample, self-guided setting, but LTPO
optimizes additional intermediate reasoning latents using policy gradients, whereas
we directly modify the given input---pixels for vision models and the token
embeddings of the observed context for language models---using gradients derived
from the likely output classes.

\paragraph{Optimizing inputs instead of model parameters.}
Optimizing the input of a fixed neural network has a long history, although mostly
for objectives different from improving ordinary test-time prediction.
\citet{simonyan14deep} optimize image pixels to maximize a class score in order to
visualize what a fixed classifier has learned. \citet{mahendran15inverting} recover
images by optimizing pixels so that a fixed network produces a target internal
representation. Adversarial-example methods likewise optimize perturbations of the
input while keeping the model fixed, but deliberately seek changes that induce
prediction errors \citep{szegedy14intriguing}. These works establish input
optimization as a general mechanism for interrogating or controlling fixed neural
networks. Our use is different: the input update is performed on the given test
sample itself, without a target label, with the aim of improving its prediction
using only the model's own likely classes. Relatedly, \citet{mum21te} learn an input
transformation at test time for distribution-shift adaptation; unlike our method,
the optimized object is a transformation module rather than the current sample
representation itself.

\paragraph{Other connections.}
Our more general theory relates to Neural Tangent Kernel theory \cite{jacot2018ntk}. Our input-gradient matrix \(K^X=J_XJ_X^\top\) is an analogous Jacobian Gram matrix, whereas \(K^\Theta\) is the same-input, output-by-output block of the empirical neural tangent kernel. Applying the top-two objective to either matrix gives the local margin velocity \(K_{aa}-K_{bb}\), making the comparison between input and parameter optimization mathematically explicit. The tendency of that diagonal difference to favor the true class is an empirical property rather than a consequence of NTK theory.

Gradient norms have been evaluated in multiple applications, e.g., to assess uncertainty of predictions \cite{ob18unc}. 

\cite{che20spur} showed that entropy minimization can reduce reliance on spurious features. We said that doFo reduces spurious classes, which likely rely on spurious features.


\section{Additional Results}
This is the appendix containing additional results. We noted that for individual model,dataset pairs there can be minor differences in the baseline accuracies. These appear to stem from using different hardware and drivers, i.e., NVidea's 4090, H100 and A6000 GPUs supported the computatoin with slightly different Python and PyTorch versions. 


\begin{figure}[H]
    \centering
    \includegraphics[width=\textwidth]{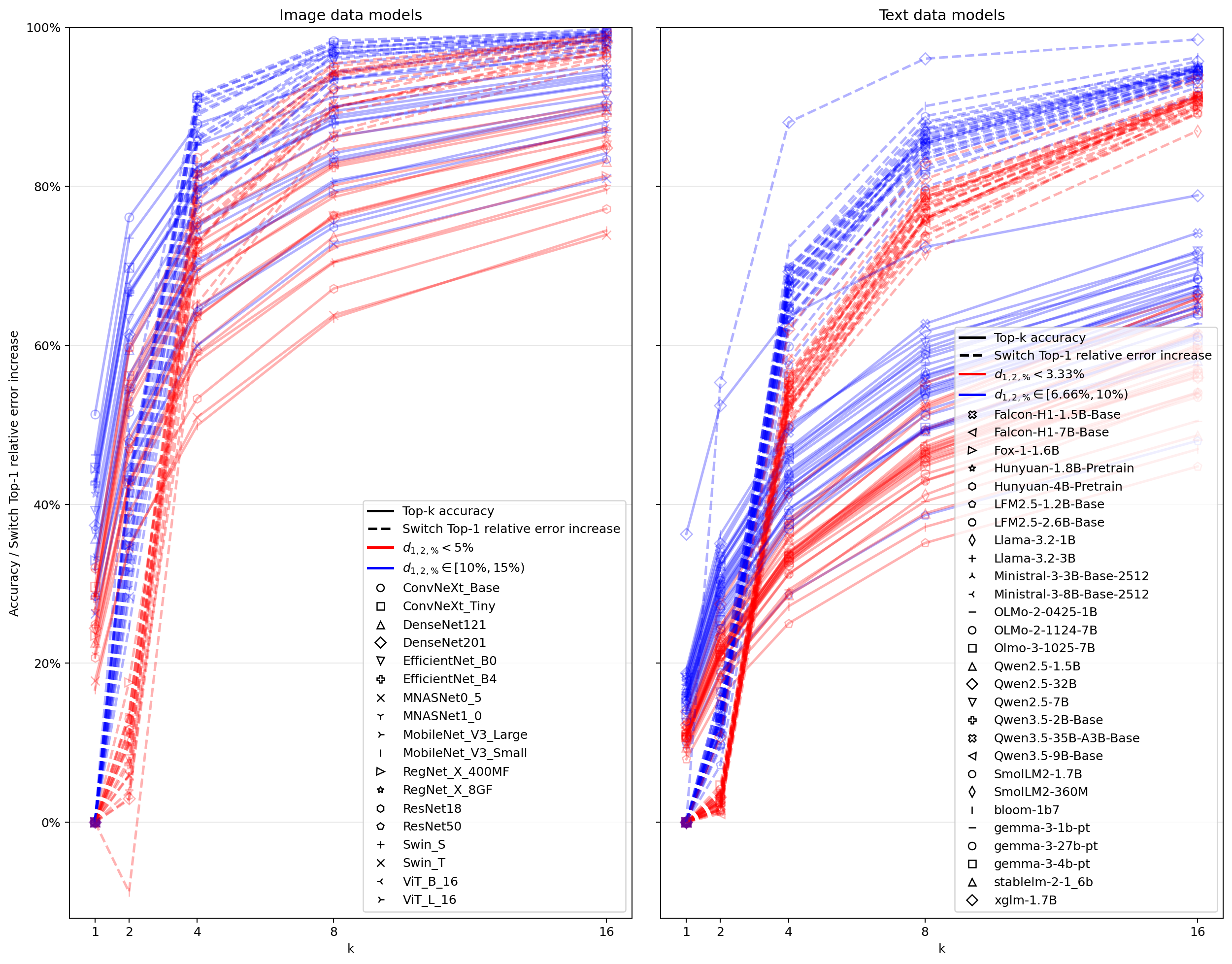}
    \caption{Per-model Top-$k$ behavior for different uncertainty regimes defined by $d_{1,2}$.}
    \label{fig:topk-d12}
\end{figure}

\begin{figure*}[t]
    \centering
    \includegraphics[width=\textwidth]{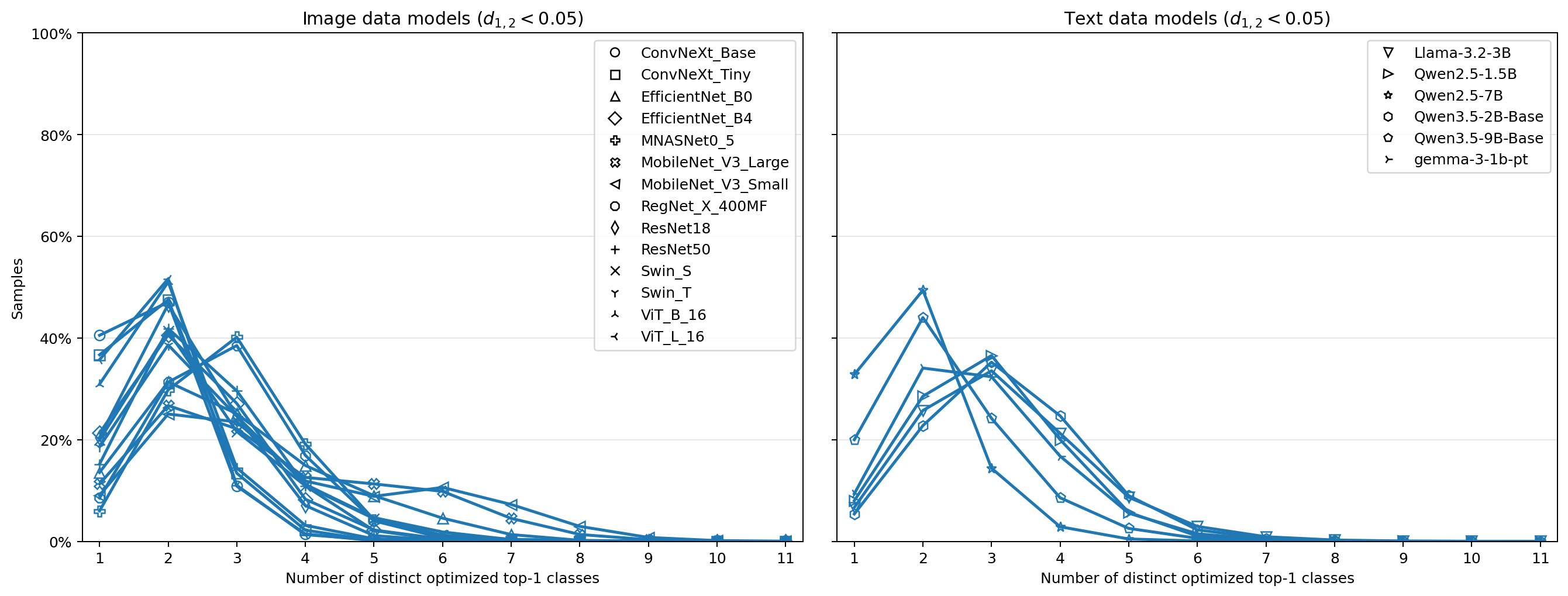}
    \caption{Per-model distribution of the number of distinct optimized Top-1 classes.}
    \label{fig:distCount}
\end{figure*}

\begin{figure*}[t]
    \centering
    \includegraphics[width=\textwidth]{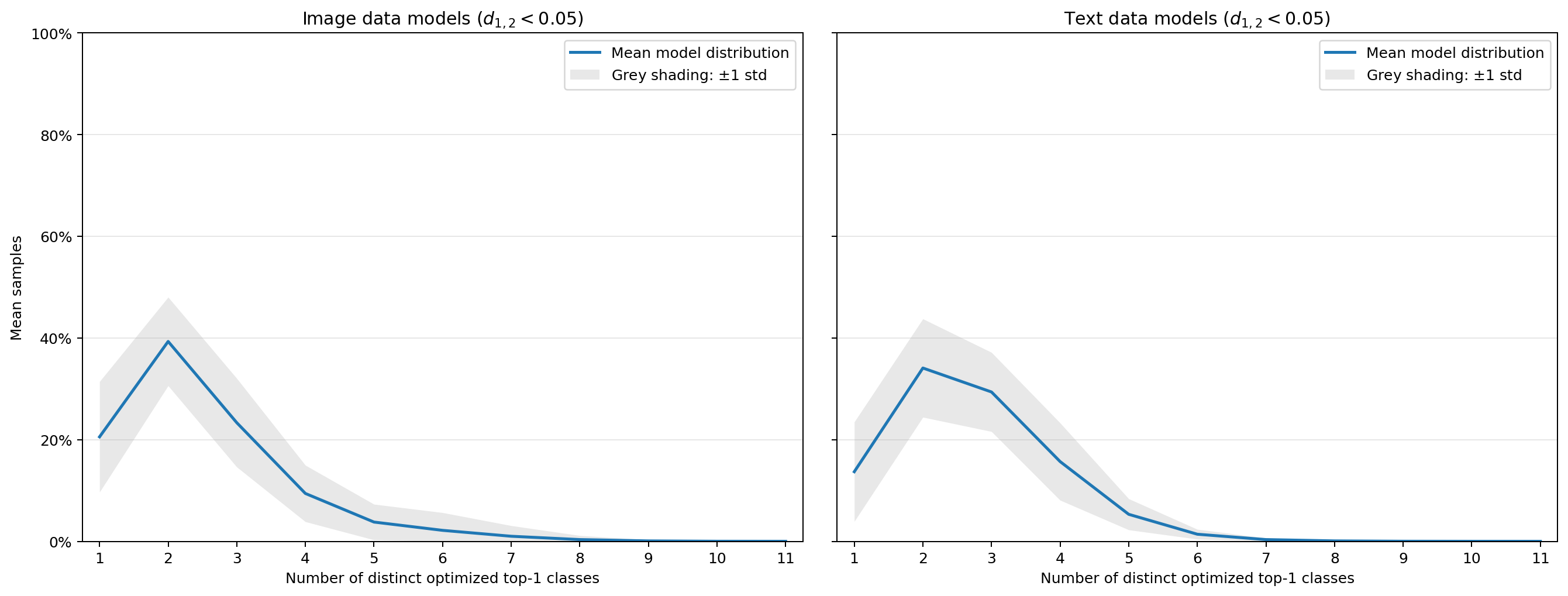}
    \caption{Per-model distribution of the number of distinct optimized Top-1 classes.}
    \label{fig:distCountAggr}
\end{figure*}

\begin{figure}[H]
    \centering
    \includegraphics[width=\textwidth]{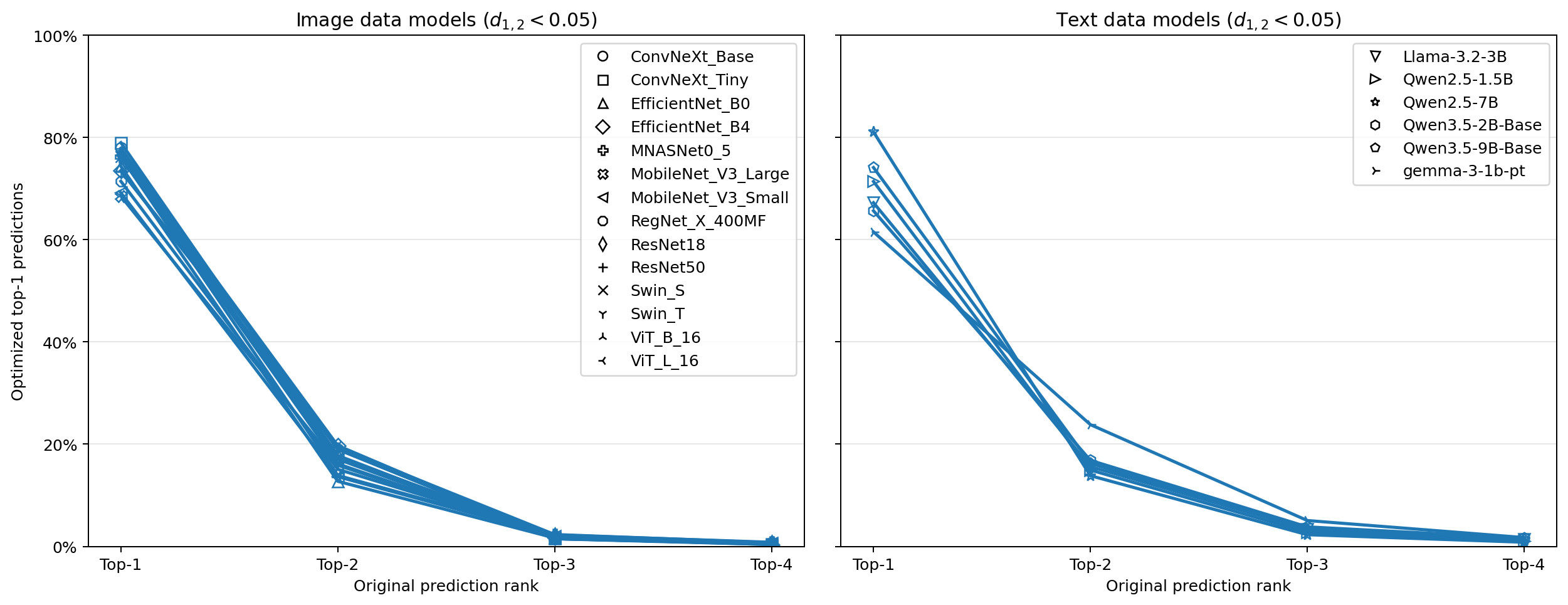}
    \caption{Per-model frequency with which optimized Top-1 predictions correspond to the original Top-1 through Top-4 predictions.}
    \label{fig:distr-topk}
\end{figure}


\begin{figure}[H]
    \centering
    \includegraphics[width=\textwidth]{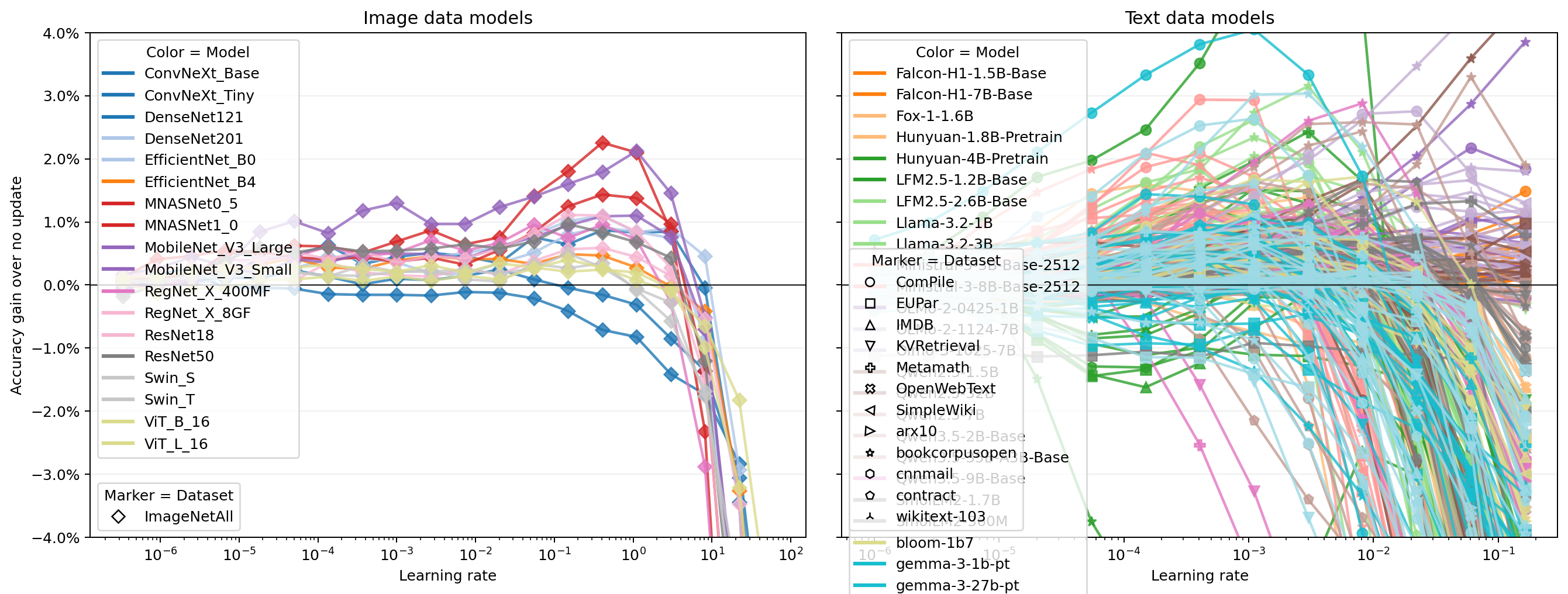}
    \caption{Per-model and per-dataset accuracy gain as a function of the learning rate.}
    \label{fig:lrall}
\end{figure}

\section{More details on experimental setup} \label{sec:experDet}

\paragraph{Motivation for SGD rather than Adam and no weight decay.}
We use SGD and not Adam, as Adam relies on moments that anticipate future or non-local changes—effectively hinting at how parameter updates might evolve when moving in a certain direction across iterations. Two properties of our method make this less relevant. First, we chose to use fixed focus classes for a single sample throughout the optimization. Second, we use mostly just a single step for stability and computational reasons. Thus, we expect limited variation in the optimization direction across iterations and anticipate that upcoming iterations provide little benefit. Following similar reasoning, we set momentum to 0, since momentum primarily prevents zig-zagging and escaping local minima by averaging gradients across multiple batches and iterations. We also set weight decay to 0, as weight decay penalizes large weights, causing the network to rely on many features. However, our philosophy is the opposite: We aim to focus on fewer relevant classes and features. Therefore, weight decay might counteract our method.

\paragraph{Hardware and Software.}
All experiments were conducted on an Ubuntu 22.04 system equipped with Python $3.12+$, PyTorch $2.5+$, CUDA $12.8+$, and multiple NVIDIA H100s, multiple A6000, and an NVIDIA RTX 4090 GPU. 

\paragraph{Datasets and Models.} 
Images: For image classification, we used Pytorch's torchvision models  (\url{https://docs.pytorch.org/vision/stable/models}) trained on ImageNet1KV1~\citep{den09} (if available). We used IMAGENET1KV1 versions as they are most widely supported among models.
For applying our method, we relied on the standard validation data of 50000 samples of ImageNet1K and added the 10k frequency matched samples from \cite{rec19Img} (often referred to Imagenet1KV2).  

Text: 
For language modeling, we use (from HuggingFace \url{https://huggingface.co/}) and the paper lists there HuggingFace name as reference.

We evaluate on datasets mostly from huggingface namely OpenWebText and open-source replication of the WebText dataset from OpenAI~\citep{ope24} (Skylion007/openwebtext), SimpleWiki, a collection Wikipedia articles written in simple English~\citep{sim24} (from \url{https://dumps.wikimedia.org/simplewiki/20250501/}), legal contracts~\citep{alb24} (albertvillanova/legal\_contracts),  Arxiv-10 (effectiveML/ArXiv-10)  ~\citep{farh22},  bookcorpus \citep{bookc15} (rojagtap/bookcorpus), CNN DailyMail \citep{cnnd17} (abisee/cnn\_dailymail Config 3.0.0), IMDB, a corpus of movie reviews introduced for sentiment analysis~\citep{imdb11} (stanfordnlp/imdb); Europarl, a multilingual parallel corpus derived from proceedings of the European Parliament~\citep{euro05}, for which we use the English side of the German--English Europarl v7 corpus; WikiText-103, a language-modeling corpus containing over 100 million tokens from verified Wikipedia articles~\citep{wiki17} (Salesforce/wikitext, configuration wikitext-103-raw-v1); ComPile, a large corpus of LLVM intermediate representations obtained from production-source programs~\citep{comp23} (llvm-ml/ComPile); Metamath, a collection of formally verified mathematical statements and proofs~\citep{metad26,meta19} (phanerozoic/Metamath); and the KV-Retrieval task from $\infty$Bench~\citep{inf24} (xinrongzhang2022/InfiniteBench), a synthetic long-context retrieval task in which a model must retrieve the value associated with a specified key.

\section{Class counts when optimizing inputs}
We (mostly) fine-tune inputs, i.e., embeddings for tokens in LLMs and pixels of images for image classification. We see that as we optimize towards the two most likely classes using different learning rates, we see that the sample's prediction alternates between the two. For very small changes, the originally predicted class does not change. For very large changes, the prediction might deviate from the top-k classes of the original sample. For the learning rates in-between, we see an interesting alternating behavior between the two most likely classes with some preference for the second most likely class. The alternating behavior within a few class predictions when changing learning rates is also empirically confirmed (see Figures \ref{fig:distCount} and \ref{fig:distCountAggr}), disproving the conjecture \cite{wan20tent} that outcomes are trivial, simply amplifying the most likely prediction. For images (left panel) and LLMs (right panel), 80\% of samples lead to 3 or less, respectively, 4 or less distinct samples, despite applying more than 10 learning rates. Thus, most alterations are between few classes.

\section{General theory for input tuning}\label{app:general}

In our implementation only the input is optimized: the pixels of an image, or the token embeddings of the context for a language model; all weights stay fixed. This appendix analyses this for an arbitrary differentiable network. The toy model in the main manuscript is exactly this situation for a one-layer feature extractor.

\subsection{Setting}

Let $f(x)\in\mathbb{R}^{|C|}$ be the logits of the fixed network for input $x\in\mathbb{R}^d$ (for a language model, $x$ is the concatenation of the token embeddings of the context). Write $\nabla_xf_c$ for the saliency map of class $c$,
\begin{equation}\label{eq:app-K}
s_c:=\|\nabla_xf_c\|^2\quad\text{(the \emph{input sensitivity} of class $c$)},\qquad
K_{ab}:=\langle\nabla_xf_a,\nabla_xf_b\rangle\quad(K_{aa}=s_a),
\end{equation}
so that $K=JJ^{\top}$ with $J=\partial f/\partial x$ is the Gram matrix of the saliency maps. For an input coordinate $i$ (a pixel, an embedding dimension) and classes $a,b$, coordinate $i$ is \emph{shared} if $\partial_if_a\,\partial_if_b\neq0$, \emph{non-contrastive} if $\partial_if_a\,\partial_if_b>0$, \emph{contrastive} if $\partial_if_a\,\partial_if_b<0$, and \emph{$a$-specific} if $\partial_if_b=0\neq\partial_if_a$. Thus $K_{ab}=\sum_i\partial_if_a\,\partial_if_b$ is the non-contrastive minus the contrastive sharing mass between $a$ and $b$, and $s_a=\sum_i(\partial_if_a)^2$ sums the squared sensitivities of class $a$ over all input features, shared and specific.

All methods minimize a loss $L$ that depends on $x$ only through the logits, by gradient descent on the input with the weights $\tilde p_c$ treated as constants. Their logit-gradients $g:=\partial L/\partial f$ are, with $p=\mathrm{softmax}(f)$, $\bar f=\sum_cp_cf_c$, focus set $F$ and weights $w_c\ge0$ supported on $F$ ($w_c=\tilde p_c$, or $w_c=1$ for the unweighted variant),
\begin{equation}\label{eq:app-grads}
\text{iFo: } g=-w;\qquad
\text{single class }k{=}1\text{ (target }a\text{): } g=-e_a;\qquad
\text{doFo: } g=\mathbf 1_{C\setminus F}/|C\setminus F|;\qquad
\text{entropy: } g_c=p_c(\bar f-f_c).
\end{equation}
The entropy logit-gradient is a contrast, $\sum_cg_c=0$, whose per-class weight $p_c|f_c-\bar f|$ vanishes for the focus classes of an uncertain sample; the iFo logit-gradient is a constant. (PASLE's loss $-\log\sum_{c\in F}p_c$ has $g_c=-p_c(1-p_F)/p_F$ on $F$ and $g_c=p_c$ off $F$: weighted iFo scaled by $(1-p_F)/p_F$ plus $p_c$-weighted doFo, both vanishing when the mass is in $F$.)

\subsection{Margin dynamics (exact for any network)}

\begin{theorem}[Margin dynamics under input tuning]\label{thm:app-flow}
Let the input follow the gradient flow $\dot x=-\nabla_xL(f(x))$ of one of the losses \eqref{eq:app-grads}, for any differentiable network. Then, at every time,
\begin{equation}\label{eq:app-flow}
\frac{d}{dt}(f_a-f_b)=-(e_a-e_b)^{\top}Kg ,
\end{equation}
that is,
\begin{align}
\text{iFo, }F=\{a,b\}:&\quad \tfrac{d}{dt}(f_a-f_b)=w_a(s_a-K_{ab})-w_b(s_b-K_{ab}),\label{eq:app-m-ifo}\\
\text{iFo, unweighted:}&\quad \tfrac{d}{dt}(f_a-f_b)=s_a-s_b,\label{eq:app-m-ifo-unw}\\
\text{iFo, general }F:&\quad \tfrac{d}{dt}(f_a-f_b)=\sum_{k\in F}w_k(K_{ak}-K_{bk}),\label{eq:app-m-ifoF}\\
k{=}1\text{ (target }a):&\quad \tfrac{d}{dt}(f_a-f_b)=s_a-K_{ab},\label{eq:app-m-k1}\\
\text{doFo:}&\quad \tfrac{d}{dt}(f_a-f_b)=-\,\mathrm{mean}_{c\notin F}(K_{ac}-K_{bc}),\label{eq:app-m-dofo}\\
\text{entropy:}&\quad \tfrac{d}{dt}(f_a-f_b)=\sum_{c\in C}p_c(f_c-\bar f)(K_{ac}-K_{bc}).\label{eq:app-m-ent}
\end{align}
\end{theorem}
\begin{proof}
$\nabla_xL=J^{\top}g$, so $\dot f=J\dot x=-JJ^{\top}g=-Kg$; multiply by $(e_a-e_b)^{\top}$ and insert $g$.
\end{proof}

Equation \eqref{eq:app-m-ifo-unw} is the central identity. The margin between the two focus classes moves towards the class whose logit is more sensitive to the input, at every instant and for every network; the sharing between the two focus classes, $K_{ab}$, cancels. Sharing enters through the input sensitivities themselves: a shared feature $i$ contributes $(\partial_if_a)^2$ to $s_a$ and $(\partial_if_b)^2$ to $s_b$, so the class that uses the shared evidence more strongly gains from it. Integrated over the flow, $f_a-f_b$ increases overall iff $\int(s_a-s_b)\,dt>0$ along the path.

\begin{lemma}[One step]\label{lem:app-step}
One gradient step $x'=x-\eta\nabla_xL$ is the Euler step of the flow:
$f(x')=f(x)-\eta Kg+\rho$, where $\rho=0$ if $f$ is affine in $x$ on the segment from $x$ to $x'$, and $|\rho_c|\le\tfrac{M\eta^2}{2}\|J^{\top}g\|^2$ if the Hessians $\nabla_x^2f_c$ have operator norm at most $M$ on that segment. Hence the identities of Theorem~\ref{thm:app-flow} hold for the step with $\tfrac{d}{dt}(\cdot)$ replaced by $\Delta(\cdot)/\eta$, exactly for a linear classifier and for piecewise-linear networks (ReLU, leaky ReLU, max-pooling) as long as the step does not change the activation pattern of the sample, and to first order in $\eta$ otherwise (e.g.\ token embeddings of a transformer).
\end{lemma}
\begin{proof}
Taylor's theorem with $x'-x=-\eta J^{\top}g$; on a region where $f$ is affine in $x$ the remainder vanishes.
\end{proof}

\begin{lemma}[To overtake, the second class must be the more sensitive one]\label{lem:app-second}
Let $a$ be the most likely and $b$ the second most likely class, $p_a\ge p_b$, and $F=\{a,b\}$ with weights $w=(p_a,p_b)$ or $w=(1,1)$. Then the margin $f_a-f_b$ decreases under iFo only if $s_b>s_a$.
\end{lemma}
\begin{proof}
By \eqref{eq:app-m-ifo}, $\tfrac{d}{dt}(f_a-f_b)<0$ means $p_b(s_b-K_{ab})>p_a(s_a-K_{ab})$. If $s_a\ge K_{ab}$, the right-hand side is at least $p_b(s_a-K_{ab})$, hence $s_b>s_a$. If $s_a<K_{ab}$, then $K_{ab}\le\sqrt{s_as_b}<\sqrt{K_{ab}s_b}$ gives $K_{ab}<s_b$, hence $s_b>K_{ab}>s_a$.
\end{proof}
So with two focus classes there is exactly one way in which iFo can correct a prediction: the true class, currently second, must be the more input-sensitive one. This is not an additional assumption of the analysis but a property of the method. It is also the condition of Theorem 2 (for toy example in the main paper): there $s_0=c_0^2+c_4^2$, $s_1=c_1^2+c_5^2$, and for $\delta_4>\delta_1$ the condition $\delta_4+\delta_1>-2$ is equivalent to $(1+\delta_4)^2>(1+\delta_1)^2$, i.e.\ to $s_0>s_1$.

\begin{remark}[More than two focus classes]\label{rem:app-k3}
For $|F|\ge3$, \eqref{eq:app-m-ifoF} gives, unweighted,
\[
\tfrac{d}{dt}(f_a-f_b)=(s_a-s_b)+\sum_{c\in F\setminus\{a,b\}}(K_{ac}-K_{bc}):
\]
a class can now overtake \emph{without} being the more sensitive one, namely if it shares more (non-contrastively) with the other focus classes than its competitor does. With $n_f=|F|=2$, as used in our experiments, this channel is absent, and the sharing between the two focus classes is neutral.
\end{remark}

\begin{corollary}[Self-confirmation: the Tent conjecture]\label{cor:app-confirm}
\leavevmode
\begin{enumerate}
\item[(a)] Single class: $\tfrac{d}{dt}(f_a-f_b)=s_a-K_{ab}\ge\sqrt{s_a}\big(\sqrt{s_a}-\sqrt{s_b}\big)$. If $s_a\ge s_b$ for all $b$, optimizing the most likely class alone can only confirm it.
\item[(b)] Entropy with the mass on two classes ($p_a+p_b=1$): with $m=f_a-f_b$,
\begin{equation}\label{eq:app-ent2}
\dot m=p_ap_b\,m\,\|\nabla_xf_a-\nabla_xf_b\|^2 ,
\end{equation}
so $m(t)=m(0)\exp\!\big(\int_0^tp_ap_b\|\nabla_xm\|^2\big)$: the sign of the margin never changes, $|m|$ grows, and a tie is stationary. Entropy minimization cannot correct an uncertain prediction; it amplifies the observed margin, noise included.
\item[(c)] Unweighted iFo overturns the prediction in one step iff $\mathrm{sgn}(s_a-s_b)\neq\mathrm{sgn}(f_a-f_b)$ and $\eta>|f_a-f_b|/|s_a-s_b|$; the required step is smallest for uncertain samples.
\end{enumerate}
\end{corollary}
\begin{proof}
(a) Cauchy--Schwarz. (b) $f_a-\bar f=p_b\,m$ and $f_b-\bar f=-p_a\,m$, so $-\nabla_xH=p_ap_bm\,\nabla_xm$ and $\dot m=\langle\nabla_xm,\dot x\rangle$. (c) is \eqref{eq:app-m-ifo-unw} with Lemma~\ref{lem:app-step}.
\end{proof}

\begin{proposition}[Probability-weighted iFo]\label{prop:app-weighted}
With $w=(p_a,p_b)$, $q=p_a+p_b$ and $m=f_a-f_b$,
\begin{equation}\label{eq:app-weighted}
\tfrac{d}{dt}(f_a-f_b)=\frac q2\Big[(s_a-s_b)+\tanh(m/2)\,\|\nabla_xf_a-\nabla_xf_b\|^2\Big].
\end{equation}
The first term is the unweighted dynamics at half speed; the second has the sign of the observed margin, vanishes at a tie, and is the entropy dynamics \eqref{eq:app-ent2} to leading order in $m$ for $q=1$. On the uncertain set $|m|\le\tau$ it is bounded by $\tfrac{q\tau}{4}\|\nabla_xf_a-\nabla_xf_b\|^2$.
\end{proposition}
\begin{proof}
$p_a=\tfrac q2+t$, $p_b=\tfrac q2-t$ with $t=\tfrac q2\tanh\tfrac m2$ (since $p_a/p_b=e^m$); insert into \eqref{eq:app-m-ifo}.
\end{proof}

\begin{theorem}[Input-feature amplification; general form of Theorem 1 (for toy example in the main paper)]\label{thm:app-amp}
For $F=\{a,b\}$, the iFo step and the single-class step change input coordinate $i$ by $\Delta^{iFo}x_i=\eta\,(w_a\partial_if_a+w_b\partial_if_b)$ and $\Delta^{(1)}x_i=\eta\,\partial_if_a$. Relative to an $a$-specific coordinate, a shared coordinate is updated by iFo by the factor $\big|1+\tfrac{w_b\,\partial_if_b}{w_a\,\partial_if_a}\big|$ more than by the single-class step; the factor exceeds $1$ iff $i$ is non-contrastive or $w_b|\partial_if_b|>2w_a|\partial_if_a|$, and is below $1$ iff $i$ is contrastive and $w_b|\partial_if_b|<2w_a|\partial_if_a|$. It does not depend on $\eta$ or on the scale of $w$.
\end{theorem}
\begin{proof}
$\nabla_xL^{iFo}=-\sum_{k\in F}w_k\nabla_xf_k$ coordinate-wise; divide the shared by the $a$-specific update and use $|1+t|>1\iff t>0$ or $t<-2$.
\end{proof}

\subsection{Input noise: general form of Theorem 2 (for toy example in the main paper)}

\begin{theorem}[iFo is a deterministic shift of a noisy margin]\label{thm:app-noise}
Let the observed input be $x=x^\ast+\varepsilon$ with $\varepsilon$ small enough that $J$ is the same at $x$ and $x^\ast$ (the affine regime of Lemma~\ref{lem:app-step}), and let the weights $w$ not depend on $\varepsilon$. Write $d:=\nabla_xf_a-\nabla_xf_b$, $m^\ast:=f_a(x^\ast)-f_b(x^\ast)$ (noise-free margin) and $\xi:=d\cdot\varepsilon$ (margin noise), so that $f_a(x)-f_b(x)=m^\ast+\xi$. One iFo step with $F=\{a,b\}$ gives
\begin{equation}\label{eq:app-shift}
f_a(x')-f_b(x')=m^\ast+\xi+\eta\,\sigma_w,\qquad \sigma_w:=w_a(s_a-K_{ab})-w_b(s_b-K_{ab})\ \ (=s_a-s_b\text{ unweighted}),
\end{equation}
i.e.\ the noise passes through unchanged and a noise-independent shift is added. Consequently, for the true class $y\in\{a,b\}$ and any noise distribution, the error probability of the decision between $a$ and $b$ decreases iff the shift of the margin $f_y-f_j$ is positive; for unweighted iFo this is iff the true class is the more input-sensitive one, and with $\alpha:=m^\ast/(s_a-s_b)>0$,
\begin{equation}\label{eq:app-alpha}
f_a(x')-f_b(x')=\Big(1+\frac\eta\alpha\Big)m^\ast+\xi .
\end{equation}
\end{theorem}
\begin{proof}
$\Delta f=\eta Kw$ with $K$, $w$ independent of $\varepsilon$; the error statement is the monotonicity of the c.d.f.\ of $\xi$; \eqref{eq:app-alpha} is \eqref{eq:app-m-ifo-unw} divided by $m^\ast$.
\end{proof}

{\tiny
\setlength{\tabcolsep}{1pt}
\begingroup
\scriptsize
\begin{longtable}{llrrrrrrrr}
\caption{Policy accuracy improvements over the original model baseline by model and method, including aggregates across all models and by data modality. Reported p-values compare policies with fixed lr and are Holm-adjusted across all four MaxSum policies and displayed methods within each model or aggregate group.}
\label{tab:statTabPolicy} \\
\toprule
Model & method & \shortstack{Baseline\\$Acc_{Te,1,2,\%}$} & \shortstack{fixed lr\\$\Delta_{Acc}$} & \shortstack{MaxSum4\\$\Delta_{Acc}$} & \shortstack{MaxSum16\\$\Delta_{Acc}$} & \shortstack{MaxSum-\\Prune4 $\Delta_{Acc}$} & \shortstack{MaxSum-\\Prune16 $\Delta_{Acc}$} & \shortstack{MaxSum16\\p} & \shortstack{MaxSum-\\Prune16 p} \\
\midrule
\endfirsthead
\toprule
Model & method & \shortstack{Baseline\\$Acc_{Te,1,2,\%}$} & \shortstack{fixed lr\\$\Delta_{Acc}$} & \shortstack{MaxSum4\\$\Delta_{Acc}$} & \shortstack{MaxSum16\\$\Delta_{Acc}$} & \shortstack{MaxSum-\\Prune4 $\Delta_{Acc}$} & \shortstack{MaxSum-\\Prune16 $\Delta_{Acc}$} & \shortstack{MaxSum16\\p} & \shortstack{MaxSum-\\Prune16 p} \\
\midrule
\endhead
All image & Entropy & 31.632\% & +0.757 & +0.786 & +0.794 & +0.837 & +0.870 & 1.0000 & 1.0000 \\
All image & iFo & 31.634\% & +0.656 & +0.728 & +0.736 & +0.725 & +0.733 & 1.0000 & 1.0000 \\
All image & doFo & 31.631\% & +0.139 & -0.157 & -0.160 & +0.044 & +0.010 & 1.0000 & 1.0000 \\
All image & $iFo^2$ & 31.627\% & +0.752 & +0.629 & +0.632 & +0.736 & +0.734 & 1.0000 & 1.0000 \\
All text & Entropy & 13.163\% & +0.588 & +0.688 & +0.742 & +0.761 & +0.830 & 8.00e-05* & 8.00e-05* \\
All text & iFo & 13.172\% & +0.524 & +0.602 & +0.642 & +0.704 & +0.759 & 8.00e-05* & 8.00e-05* \\
All text & doFo & 13.897\% & +0.152 & +0.122 & +0.096 & +0.313 & +0.352 & 1.0000 & 8.00e-05* \\
All text & $iFo^2$ & 15.263\% & +0.389 & +0.535 & +0.563 & +0.615 & +0.647 & 8.00e-05* & 8.00e-05* \\
\bottomrule
\end{longtable}
\endgroup

\begingroup
\scriptsize
\begin{longtable}{llrrrrrrrr}
\caption{Policy accuracy improvements over the original model baseline by dataset and method, averaging matched models equally within each dataset. Reported p-values compare policies with fixed lr and are Holm-adjusted across all four MaxSum policies and displayed methods within each dataset.}
\label{tab:statTabPolicyData} \\
\toprule
Dataset & method & \shortstack{Baseline\\$Acc_{Te,1,2,\%}$} & \shortstack{fixed lr\\$\Delta_{Acc}$} & \shortstack{MaxSum4\\$\Delta_{Acc}$} & \shortstack{MaxSum16\\$\Delta_{Acc}$} & \shortstack{MaxSum-\\Prune4 $\Delta_{Acc}$} & \shortstack{MaxSum-\\Prune16 $\Delta_{Acc}$} & \shortstack{MaxSum16\\p} & \shortstack{MaxSum-\\Prune16 p} \\
\midrule
\endfirsthead
\toprule
Dataset & method & \shortstack{Baseline\\$Acc_{Te,1,2,\%}$} & \shortstack{fixed lr\\$\Delta_{Acc}$} & \shortstack{MaxSum4\\$\Delta_{Acc}$} & \shortstack{MaxSum16\\$\Delta_{Acc}$} & \shortstack{MaxSum-\\Prune4 $\Delta_{Acc}$} & \shortstack{MaxSum-\\Prune16 $\Delta_{Acc}$} & \shortstack{MaxSum16\\p} & \shortstack{MaxSum-\\Prune16 p} \\
\midrule
\endhead
arx10 & Entropy & 13.625\% & +0.445 & +0.507 & +0.531 & +0.542 & +0.585 & 0.4098 & 0.0080* \\
arx10 & iFo & 13.679\% & +0.473 & +0.493 & +0.491 & +0.564 & +0.593 & 1.0000 & 0.2028 \\
arx10 & doFo & 15.021\% & -0.043 & -0.100 & -0.152 & +0.110 & +0.115 & 1.0000 & 0.1399 \\
arx10 & $iFo^2$ & 17.876\% & +0.386 & +0.454 & +0.463 & +0.505 & +0.523 & 0.4501 & 0.0028* \\
bookcorpusopen & Entropy & 11.183\% & +1.113 & +1.029 & +1.200 & +1.265 & +1.412 & 1.0000 & 0.1114 \\
bookcorpusopen & iFo & 11.183\% & +0.603 & +0.888 & +1.017 & +1.130 & +1.258 & 0.0455* & 0.0022* \\
bookcorpusopen & doFo & 11.264\% & +0.422 & +0.378 & +0.421 & +0.722 & +0.809 & 1.0000 & 0.0118* \\
bookcorpusopen & $iFo^2$ & 12.347\% & +0.470 & +0.711 & +0.794 & +0.921 & +1.001 & 0.4217 & 4.00e-04* \\
\bottomrule
\end{longtable}
\endgroup

\begingroup
\scriptsize
\begin{longtable}{lllrrrrrrrr}
\caption{Policy accuracy improvements over the original model baseline for each individual model/dataset pair and method. Each row has one test unit, so p-values are unavailable (--). Reported p-values compare policies with fixed lr and are Holm-adjusted across all four MaxSum policies and displayed methods within each model/dataset pair.}
\label{tab:statTabPolicyAll} \\
\toprule
Model & Dataset & method & \shortstack{Baseline\\$Acc_{Te,1,2,\%}$} & \shortstack{fixed lr\\$\Delta_{Acc}$} & \shortstack{MaxSum4\\$\Delta_{Acc}$} & \shortstack{MaxSum16\\$\Delta_{Acc}$} & \shortstack{MaxSum-\\Prune4 $\Delta_{Acc}$} & \shortstack{MaxSum-\\Prune16 $\Delta_{Acc}$} & \shortstack{MaxSum16\\p} & \shortstack{MaxSum-\\Prune16 p} \\
\midrule
\endfirsthead
\toprule
Model & Dataset & method & \shortstack{Baseline\\$Acc_{Te,1,2,\%}$} & \shortstack{fixed lr\\$\Delta_{Acc}$} & \shortstack{MaxSum4\\$\Delta_{Acc}$} & \shortstack{MaxSum16\\$\Delta_{Acc}$} & \shortstack{MaxSum-\\Prune4 $\Delta_{Acc}$} & \shortstack{MaxSum-\\Prune16 $\Delta_{Acc}$} & \shortstack{MaxSum16\\p} & \shortstack{MaxSum-\\Prune16 p} \\
\midrule
\endhead
bloom-1b7 & arx10 & Entropy & 14.640\% & +0.850 & +0.683 & +0.760 & +0.833 & +0.910 & -- & -- \\
bloom-1b7 & arx10 & iFo & 14.640\% & +0.703 & +0.697 & +0.733 & +0.670 & +0.763 & -- & -- \\
bloom-1b7 & arx10 & doFo & 14.737\% & +0.020 & -0.140 & -0.180 & +0.070 & +0.160 & -- & -- \\
bloom-1b7 & arx10 & $iFo^2$ & 14.640\% & +0.687 & +0.640 & +0.687 & +0.620 & +0.647 & -- & -- \\
bloom-1b7 & bookcorpusopen & Entropy & 11.663\% & +1.630 & +1.087 & +1.153 & +1.400 & +1.467 & -- & -- \\
bloom-1b7 & bookcorpusopen & iFo & 11.663\% & +0.747 & +0.247 & +0.297 & +0.907 & +0.767 & -- & -- \\
bloom-1b7 & bookcorpusopen & doFo & 11.663\% & +0.037 & -0.023 & +0.003 & +0.107 & +0.180 & -- & -- \\
bloom-1b7 & bookcorpusopen & $iFo^2$ & 11.663\% & +0.853 & +0.500 & +0.517 & +0.420 & +0.463 & -- & -- \\
\bottomrule
\end{longtable}
\endgroup

}

\begin{table*}[t]
\centering
\small
\caption{Fixed learning rates by model and objective for Table 0. Each cell lists the selected rate in validation-fold order; each fold fits on its pooled training samples across datasets. Missing conditions are shown as --. Uncertain samples are the lowest baseline top-1/top-2 probability gaps within each model/dataset evaluation population (image=15\%, tabular=10\%, text=10\%), selected before LR fitting. Samples outside the cutoff retain their baseline predictions in full-population accuracy. Top logits by data modality: image=512, tabular=8192, text=8192.}
\label{tab:fixed-lr-per-model}
\resizebox{\textwidth}{!}{%
\begin{tabular}{lcccc}
\toprule
Model & Entropy & iFo & doFo & $iFo^2$ \\
\midrule
bloom-1b7 & 0.002996065, 0.002996065 & 0.0004049612, 0.002996065 & 0.0001488828, 0.0004049612 & 0.002996065, 0.002996065 \\
ConvNeXt\_Base & 2.4661333e-06, 2.4661333e-06 & 2.4661333e-06, 2.4661333e-06 & 0.054664526, 0.00099868828 & 2.4661333e-06, 2.4661333e-06 \\
ConvNeXt\_Tiny & 6.7078827e-06, 9.0666668e-07 & 0.020097252, 3.3333333e-07 & 1.1000497, 0.00036716482 & 4.9627601e-05, 9.0666668e-07 \\
DenseNet121 & 1.1000497, 0.40443003 & 2.9921353, 0.40443003 & 3.3333333e-07, 2.4661333e-06 & 0.40443003, 0.020097252 \\
DenseNet201 & 0.40443003, 0.054664526 & 0.14868751, 1.1000497 & 9.0666668e-07, 0.0027164323 & 0.14868751, 0.14868751 \\
EfficientNet\_B0 & 0.40443003, 0.40443003 & 1.1000497, 0.40443003 & 1.8245441e-05, 2.4661333e-06 & 0.14868751, 0.40443003 \\
EfficientNet\_B4 & 0.14868751, 0.40443003 & 0.14868751, 0.40443003 & 9.0666668e-07, 1.8245441e-05 & 0.14868751, 0.40443003 \\
Falcon-H1-1.5B-Base & 0.16399358, 0.16399358 & 0.16399358, 0.16399358 & 0.16399358, 0.16399358 & 0.16399358, 0.16399358 \\
Falcon-H1-7B-Base & 0.16399358, 0.16399358 & 0.16399358, 0.16399358 & 2.0123647e-05, 0.008149297 & 0.002996065, 0.16399358 \\
Fox-1-1.6B & 0.008149297, 0.008149297 & 0.008149297, 0.008149297 & 0.0001488828, 7.3984002e-06 & 0.008149297, 0.008149297 \\
gemma-3-1b-pt & 0.0004049612, 0.0004049612 & 0.0011014945, 0.0004049612 & 1e-06, 2.7200001e-06 & 0.002996065, 0.002996065 \\
gemma-3-27b-pt & 0.0011014945, 0.0011014945 & 0.0011014945, 0.0011014945 & 0.0011014945, 0.0011014945 & 0.0011014945, 0.0004049612 \\
gemma-3-4b-pt & 0.0011014945, 0.0011014945 & 0.0011014945, 0.0011014945 & 7.3984002e-06, 2.7200001e-06 & 0.0011014945, 0.0011014945 \\
Hunyuan-1.8B-Pretrain & 0.0004049612, 0.0004049612 & 0.0004049612, 0.0004049612 & 7.3984002e-06, 7.3984002e-06 & 0.0011014945, 0.0011014945 \\
Hunyuan-4B-Pretrain & 1e-06, 7.3984002e-06 & 0.0004049612, 0.0004049612 & 1e-06, 7.3984002e-06 & 5.4736323e-05, 0.0001488828 \\
LFM2.5-1.2B-Base & 0.002996065, 0.002996065 & 0.002996065, 0.002996065 & 2.7200001e-06, 2.7200001e-06 & 0.002996065, 0.002996065 \\
LFM2.5-2.6B-Base & 0.0011014945, 0.002996065 & 0.0004049612, 0.0011014945 & 0.0004049612, 0.0004049612 & 0.0011014945, 0.0011014945 \\
Llama-3.2-1B & 0.0004049612, 0.0011014945 & 0.0004049612, 0.0001488828 & 7.3984002e-06, 5.4736323e-05 & 0.0001488828, 0.0004049612 \\
Llama-3.2-3B & 0.0011014945, 0.0011014945 & 0.0004049612, 0.0011014945 & 2.0123647e-05, 5.4736323e-05 & 0.0011014945, 0.002996065 \\
Ministral-3-3B-Base-2512 & 0.0004049612, 0.0004049612 & 0.0004049612, 0.0004049612 & 5.4736323e-05, 2.7200001e-06 & 0.0004049612, 0.0004049612 \\
Ministral-3-8B-Base-2512 & 0.0004049612, 0.0004049612 & 0.0004049612, 0.0004049612 & 2.0123647e-05, 5.4736323e-05 & 0.0004049612, 0.0004049612 \\
MNASNet0\_5 & 1.1000497, 0.40443003 & 1.1000497, 0.40443003 & 0.0027164323, 4.9627601e-05 & 0.40443003, 1.1000497 \\
MNASNet1\_0 & 0.40443003, 1.1000497 & 0.14868751, 1.1000497 & 3.3333333e-07, 4.9627601e-05 & 0.14868751, 0.40443003 \\
MobileNet\_V3\_Large & 0.40443003, 1.1000497 & 1.1000497, 0.40443003 & 6.7078827e-06, 2.4661333e-06 & 0.40443003, 0.40443003 \\
MobileNet\_V3\_Small & 1.1000497, 1.1000497 & 1.1000497, 1.1000497 & 4.9627601e-05, 0.00013498707 & 0.40443003, 0.14868751 \\
OLMo-2-0425-1B & 0.060291756, 0.060291756 & 1e-06, 1e-06 & 0.022166088, 0.022166088 & 7.3984002e-06, 2.0123647e-05 \\
OLMo-2-1124-7B & 0.060291756, 0.060291756 & 0.0001488828, 5.4736323e-05 & 0.022166088, 0.022166088 & 5.4736323e-05, 7.3984002e-06 \\
Olmo-3-1025-7B & 0.022166088, 0.022166088 & 5.4736323e-05, 5.4736323e-05 & 0.008149297, 0.008149297 & 2.7200001e-06, 2.7200001e-06 \\
Qwen2.5-1.5B & 0.0011014945, 0.002996065 & 0.0011014945, 0.0004049612 & 0.0011014945, 0.0004049612 & 0.002996065, 0.002996065 \\
Qwen2.5-32B & 0.022166088, 0.16399358 & 0.060291756, 0.008149297 & 0.0001488828, 2.7200001e-06 & 0.022166088, 0.060291756 \\
Qwen2.5-7B & 0.008149297, 0.002996065 & 0.002996065, 0.0001488828 & 0.0011014945, 0.0011014945 & 0.0011014945, 0.0004049612 \\
Qwen3.5-2B-Base & 0.0004049612, 0.0001488828 & 0.0011014945, 0.0004049612 & 2.0123647e-05, 2.7200001e-06 & 0.0011014945, 0.0004049612 \\
Qwen3.5-9B-Base & 5.4736323e-05, 0.0004049612 & 0.0001488828, 0.0001488828 & 2.0123647e-05, 1e-06 & 2.7200001e-06, 2.0123647e-05 \\
RegNet\_X\_400MF & 0.054664526, 0.40443003 & 0.14868751, 0.40443003 & 6.7078827e-06, 3.3333333e-07 & 0.00036716482, 0.40443003 \\
RegNet\_X\_8GF & 0.14868751, 0.14868751 & 0.054664526, 0.40443003 & 3.3333333e-07, 9.0666668e-07 & 0.40443003, 0.020097252 \\
ResNet18 & 0.40443003, 1.8245441e-05 & 0.14868751, 1.8245441e-05 & 6.7078827e-06, 2.4661333e-06 & 0.14868751, 0.14868751 \\
ResNet50 & 0.14868751, 0.14868751 & 1.1000497, 0.020097252 & 3.3333333e-07, 3.3333333e-07 & 1.1000497, 0.40443003 \\
SmolLM2-1.7B & 0.002996065, 0.008149297 & 0.002996065, 0.002996065 & 7.3984002e-06, 2.0123647e-05 & 0.0011014945, 0.0011014945 \\
SmolLM2-360M & 0.022166088, 0.022166088 & 0.008149297, 0.002996065 & 2.0123647e-05, 2.0123647e-05 & 0.022166088, 0.022166088 \\
stablelm-2-1\_6b & 0.0004049612, 0.0004049612 & 0.0001488828, 0.0001488828 & 7.3984002e-06, 7.3984002e-06 & 0.0001488828, 0.0001488828 \\
Swin\_S & 0.054664526, 0.40443003 & 0.054664526, 9.0666668e-07 & 0.054664526, 0.14868751 & 6.7078827e-06, 6.7078827e-06 \\
Swin\_T & 0.14868751, 0.14868751 & 0.020097252, 1.8245441e-05 & 4.9627601e-05, 1.8245441e-05 & 0.14868751, 1.8245441e-05 \\
ViT\_B\_16 & 1.1000497, 6.7078827e-06 & 1.1000497, 1.8245441e-05 & 0.14868751, 0.00013498707 & 4.9627601e-05, 0.00036716482 \\
ViT\_L\_16 & 0.14868751, 0.14868751 & 0.40443003, 0.14868751 & 0.0073886956, 0.0073886956 & 0.020097252, 0.0073886956 \\
xglm-1.7B & 0.008149297, 0.008149297 & 0.0004049612, 0.0004049612 & 1e-06, 2.7200001e-06 & 0.0004049612, 0.0011014945 \\
\bottomrule
\end{tabular}
}
\end{table*}

\end{document}